\documentclass[lettersize,journal]{IEEEtran}
\usepackage{amsmath,amsfonts,amssymb,dsfont}
\usepackage{enumerate}
\usepackage{algorithmic}
\usepackage{algorithm}
\usepackage{array}
\usepackage{textcomp}
\usepackage{stfloats}
\usepackage{url}
\usepackage{verbatim}
\usepackage{graphicx}
\usepackage{cite}
\usepackage{textcomp,subfigure,epstopdf,color,amsthm}
\newtheorem{thm}{Theorem}
\newtheorem{remark}{Remark}

\newtheorem{coro}{Corollary}
\newtheorem{lemma}{Lemma}
\newtheorem{assumption}{Assumption}

\makeatletter

\begin{document}

\title{Efficient Learning for Selecting Top-$m$ Context-Dependent Designs}

\author{Gongbo Zhang, Sihua Chen, Kuihua Huang and Yijie Peng
\thanks{Gongbo Zhang and Yijie Peng are with the Department of Management Science and Information Systems, Guanghua School of Management, Peking University, Beijing 100871, China (e-mail: gongbozhang@pku.edu.cn; pengyijie@pku.edu.cn).}%
\thanks{Sihua Chen is with the School of Information Management, Jiangxi University of Finance and Economics, Nanchang 330013, China (e-mail: doriancsh@foxmail.com).}%
\thanks{Kuihua Huang is with the College of Systems Engineering, National University of Defense Technology, Changsha 410073, China (e-mail: khhuang@nudt.edu.cn).}%
\thanks{Corresponding authors: Sihua Chen and Yijie Peng.}
\thanks{The paper has been revised on: December 25, 2021; January 12, 2022; October 11, 2022; January 4, 2023; June 9, 2023}
}

\markboth{}%
{Zhang \MakeLowercase{\textit{et al.}}: Efficient Learning for Selecting Top-$m$ Context-Dependent Designs}


\maketitle

\begin{abstract}
We consider a simulation optimization problem for context-dependent decision-making, which aims to determine the top-$m$ designs for all contexts. Under a Bayesian framework, we formulate the optimal dynamic sampling decision as a stochastic dynamic programming problem and develop a sequential sampling policy to efficiently learn the performance of each design under each context. The asymptotically optimal sampling ratios are derived to attain the optimal large deviations rate of the worst-case probability of false selection. The proposed sampling policy is proved to be consistent, and its asymptotic sampling ratios are shown to be asymptotically optimal. Numerical experiments demonstrate that the proposed method improves the efficiency for selecting top-$m$ context-dependent designs.
\end{abstract}

\def\abstractname{Note to Practitioners}
\begin{abstract}
The performance of a given design may vary across different contexts, and better decision-making is possible by refining the available contextual information. We consider a context-dependent ranking and selection problem, which allows optimal selection to depend on contextual information obtained prior to decision-making. We develop a dynamic sampling scheme to efficiently learn and select the top-$m$ designs in all contexts. Numerical experiments, including a honeypot deception game problem and a medical resource allocation problem, demonstrate that the proposed sampling scheme significantly improves the efficiency of context-dependent ranking and selection for the top-$m$ designs.
\end{abstract}


\section{Introduction} \label{sec1}
\IEEEPARstart{S}{imulation} is a powerful tool for analyzing and optimizing complex stochastic systems. Recently, there has been interest in settings where a complex system can be simulated, and multiple decisions are made at an individual level based on available contextual information. We consider a simulation optimization problem of determining designs with the top-$m$ largest (smallest) means from a finite set of $k$ designs $\left(1 \le m < k, m,k\in\mathds{Z}^{+}\right)$ for all realized contexts. The mean performance of each design under each context is unknown and can only be estimated by Monte Carlo simulation within a finite simulation budget. The performance of each design depends on the context, and the top-$m$ designs are also context-dependent. For example, in the medical resource allocation of emergency departments during the COVID-19 pandemic, good allocation designs may vary with emergency patient flows in response to the evolving situation in the pandemic \cite{melman2021balancing,bovim2022simulating}. In other fields such as marketing and finance, contextual information can help companies achieve more revenues and user satisfaction by providing personalized pricing, advertisements, promotions, and webpage layouts based on user locations, demographics, and interests \cite{chen2020competitive,chen2022statistical}.


For a certain context and $m=1$, the problem reduces to selecting the best design under a single context, which is typically referred to as ranking and selection (R\&S) problem in the simulation literature. R\&S procedures efficiently allocate the simulation budget among a finite set of designs to learn the best one. The probability of correct selection (PCS) is a commonly used metric for evaluating the statistical efficiency of a sampling procedure in R\&S. The classical problem has been generalized to the selection of the top-$m$ designs \cite{chen2008efficient, zhang2015simulation, gao2016new,zhang2021asymptotically}, which provides more choices for decision-makers, so that they can incorporate personal preferences and adapt to unexpected events to make more resilient decisions. Compared to the top-$m$ selection in R\&S, the top-$m$ designs in our problem are dependent on contexts that are realized prior to decision-making. In this work, we aim to correctly identify the top-$m$ designs for all contexts. In the example of medical resource allocation, the waiting time of critical patients for an allocation design depends on the emergency patient flows, which vary during the COVID-19 pandemic. For example, the Delta variant is more transmissible than the Alpha variant, leading to an increasing demand for medical care. Some allocation designs may become unavailable during emergencies, such as equipment failure and staff redeployment, rendering the best allocation design with the minimum waiting time infeasible. Preparing the top-$m$ allocation designs for each situation of emergency patient flow would provide more flexible decision support for hospital management during the COVID-19 pandemic. Our sampling procedure aims to efficiently allocate a finite simulation budget among all design-context pairs. Incorrect selection of any top-$m$ designs for each context could lead to the failure of our goal. To measure the statistical efficiency of our sampling procedure across the entire context space, we use the worst-case PCS (${\rm{PCS}}_W$) as a metric, where the notion of worst-case refers to the context with the lowest PCS.

Sampling efficiency is a central issue in R\&S since simulation is usually computationally expensive. In our problem, there may exist a large number of possible designs and contexts, and all design-context pairs need to be considered in a sampling procedure. Thus learning efficiency is even more important than that in the classical R\&S problem. To enhance the efficiency of learning the design-context pairs, our research objective is to efficiently allocate the simulation budget to each design-context pair such that the ${\rm PCS}_{W}$ can be maximized. We propose a dynamic sampling policy to efficiently learn the performance for each design-context pair. To capture the asymptotic optimality of our sampling policy, we derive the asymptotically optimal sampling ratios by optimizing the large deviations rate of the worst-case probability of false selection (${\rm{PFS}}_{W}$), i.e., $1-{\rm{PCS}}_{W}$. The proposed sampling procedure is consistent, i.e., the top-$m$ designs for all contexts can eventually be selected  as the number of simulation budget goes to infinity, and its asymptotic sampling ratios are shown to be asymptotically optimal under normal sampling distributions. We present a rigorous analysis that asymptotically characterizes the sampling policy among design-context pairs under our proposed sampling procedure. In summary, the contributions of our work are threefold.
\begin{itemize}
    \item To the best of our knowledge, this paper is the first to consider the selection of the top-$m$ (in particular, $m \ne 1$) context-dependent designs in R\&S problems and propose a dynamic sampling policy under a stochastic control framework. The simulation budget is allocated to learn designs for all contexts, leading to a major challenge that makes the policy different from the classical R\&S procedures.
    \item We derive asymptotic optimality conditions for the top-$m$ context-dependent selection problem, which generalize the results in \cite{zhang2021asymptotically} and \cite{du2022rate}. Specifically, when $m \ne 1$ and $m \ne k-1$, the derived asymptotic optimality conditions serve only as a necessary condition for achieving asymptotically optimal sampling ratios. This poses a significant challenge in characterizing the asymptotic optimality of a sampling policy compared to \cite{du2022rate}. Moreover, the derived asymptotic optimality conditions capture the trade-off between sampling within a certain context and across different contexts, distinguishing our analyses from those in \cite{zhang2021asymptotically}.

    \item We design an efficient dynamic sampling policy called AOAmc, which is based on a value function approximation (VFA) one-step look ahead. We prove the consistency and asymptotic optimality of the proposed policy. Our theoretical contributions in these proofs include addressing the challenges associated with characterizing asymptotic optimality and balancing the sampling among both designs and contexts. These challenges set our proofs apart from those in \cite{zhang2021asymptotically}. The effectiveness of our sampling policy is demonstrated through numerical experiments involving synthetic examples and two realistic applications: a honeypot deception game problem and a medical resource allocation problem.
\end{itemize}

\subsection{Related Literature}

Existing R\&S procedures are often categorized into fixed-precision and fixed-budget procedures. See \cite{kim2006selecting,hunter2017parallel,hong2021review} for overviews. Fixed-precision procedures \cite{kim2001fully,frazier2014fully} allocate the simulation budget to guarantee a pre-specified PCS level, whereas fixed-budget procedures \cite{chen2000simulation,peng2018ranking} aim to maximize the posterior performance subject to a fixed simulation budget constraint. The optimal sampling and allocation decisions for a sequential fixed-budget sampling policy are governed by the Bellman equations of a Markov decision process \cite{peng2016dynamic,peng2018ranking}. In \cite{glynn2004large}, the asymptotic optimality conditions for the problem of selecting the best design are derived using the large deviations technique. The study demonstrates that achieving static asymptotically optimal sampling ratios leads to an optimal exponential decay of the frequentist probability of false selection for the best design. A sampling policy that achieves asymptotically optimal sampling ratios is thereby considered asymptotically optimal. Recent studies have extended the results in \cite{glynn2004large} to the selection of top-$m$ designs \cite{zhang2021asymptotically} and the selection of the best context-dependent designs \cite{du2022rate}. Several sequential fixed-budget sampling policies have been proved to be asymptotically optimal, either under normal sampling distributions \cite{zhang2021asymptotically,peng2018ranking,chen2019complete,avci2021getting,du2022rate,li2020context}, or under general sampling distributions \cite{chen2022balancing}. The proof of the asymptotic optimality of a sampling policy typically involves demonstrating that its asymptotic sampling ratios satisfy corresponding asymptotic optimality conditions.

While context-free R\&S problems have been extensively studied and well understood, the literature on contextual selection problems is still sparse. Contexts are also known as covariates, personalized information, side information, or auxiliary quantities in the literature. Existing sampling policies for the context-dependent selection problem can be categorized based on whether the context takes values from a finite set or an infinite set. In settings where the context space is finite, every design-context pair can be sampled. \cite{gao2019selecting,du2022rate} and \cite{jin2019analytics} extend the results in \cite{glynn2004large} to contextual selection problems. They propose sampling policies that sequentially balance the conditions satisfied by the asymptotically optimal sampling ratios. From a frequentist perspective, \cite{shen2017ranking} and \cite{shen2021ranking} derive a two-stage algorithm based on the indifference zone formulation. They utilize a linear metamodel to characterize the linear relationship between each design and the contexts. Following the pioneer work in \cite{shen2017ranking}, \cite{li2018data} considers high-dimensional contexts and generalizes the setting to a larger class of basis functions beyond linearity. In \cite{li2020context}, a Gaussian mixture model is used to group designs and contexts into clusters, ensuring that performances of design-context pairs are similar within each cluster. The paper proposes a dynamic sampling policy that achieves the asymptotic optimality conditions defined in \cite{du2022rate}. Instead of using a Gaussian mixture metamodel, \cite{cakmak2022contextual} proposes a sampling policy for a Gaussian process metamodel to maximize the posterior PCS. This policy sequentially balances the asymptotic optimality conditions similar to those in \cite{du2022rate}. In settings where the context is infinite, sampling every possible context value becomes infeasible. \cite{hu2017sequential,pearce2018continuous} and \cite{ding2019knowledge} consider the problem from a Bayesian perspective. They employ acquisition functions such as expected improvement and knowledge gradient, which are myopic approximations of the optimal sampling strategy, and derive sequential sampling policies to maximize the posterior probability of good selection under a fixed simulation budget constraint. Notably, none of the existing work considers the top-$m$ context-dependent selection problem.

Our problem is related to the literature on contextual multi-armed bandit (MAB) in machine learning. The contextual MAB problem often assumes certain structures, such as the Lipschitz structure \cite{hazan2007online,krishnamurthy2019contextual}, where expected rewards exhibit Lipschitz continuity with respect to contexts, the linear structure \cite{li2010contextual,abbasi2011improved}, where the expected reward has a linear dependency on contexts, and more general mappings from contexts to designs using neural networks \cite{allesiardo2014neural} and random forest \cite{feraud2016random}. The context-free top-$m$ selection in R\&S is similar to a pure-exploration version of the MAB problem known as the best-$m$-arm identification. Existing algorithms for best-$m$-arm identification can be categorized into fixed-confidence \cite{kalyanakrishnan2012pac,kaufmann2013information,chen2017nearly} and fixed-budget \cite{bubeck2013multiple} approaches. However, there are relatively few studies on contextual best-$m$-arm identification. To the best of our knowledge, \cite{reda2021top} is the first research that considers the best-$m$-arm identification problem with linear dependence of rewards on contexts under a fixed-confidence setting. None of the existing work on MAB focuses on the selection of top-$m$ context-dependent designs under a fixed budget setting.

The worst-case performance metric used in our problem takes into account the risk of extreme events against attacks. Optimizing this metric is related to the literature on robust optimization (RO). Robust simulation involves conducting worst-case analysis on a simulation model with unknown input distributions within an ambiguity set \cite{lam2018sensitivity,ghosh2019robust}. In RO, the parameters of the unknown input distributions are treated as decision variables. In contrast, our work focuses on optimizing design variables based on a worst-case performance metric. The context space can be associated with an ambiguity set that contains a finite number of parameters representing unknown input distributions, including the true input distribution. In light of this, our problem setting is related to the distributionally robust optimization (DRO) in the R\&S literature. DRO aims to find robust designs with the best worst-case performance \cite{gao2017robust,fan2020distributionally} or the top-$m$ worst-case performances \cite{xiao2018simulation} over the ambiguity set, where the robust designs with the worst-case performances are universal. In contrast, our problem involves the top-$m$ designs that are context-dependent, and the input distribution is fixed.

The rest of the paper is organized as follows. In Section \ref{sec2}, we formulate the studied problem. Section \ref{sec3} proposes a computationally efficient dynamic sampling procedure. In Section \ref{sec4}, we derive the asymptotically optimal sampling ratios and establish the asymptotic optimality of the proposed sampling procedure. Section \ref{sec5} presents numerical examples and computational results, and Section \ref{sec6} concludes the paper and outlines future directions.

\section{Problem Formulation} \label{sec2}

\begin{figure}[!h]
\centering
\includegraphics[width=0.37\textwidth]{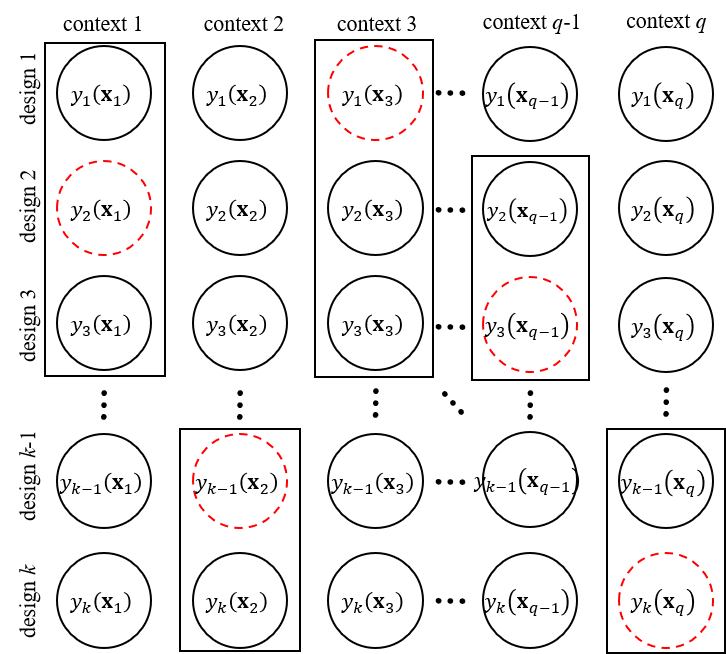}
\caption{Illustration of selecting top-$m$ designs for all contexts. The red dotted circle denotes the best design, and the black solid line contains the top-$m$ designs in each context, where the number of $m$ can be different among contexts.}
\label{figconill}
\end{figure}

Suppose there are $k$ competing designs and the performance of each design $y _{i}\left( \boldsymbol x\right) \in \mathbb{R}$, $i=1,\cdots,k$ depends on a vector of context $ {\boldsymbol x} = \left(x_{1},\cdots,x_{d}\right)^{T}$ for ${\boldsymbol x} \in \mathcal{X}\subseteq {\mathbb{R}^d}$, where $y _{i}\left( {\boldsymbol x}\right)$ is unknown and can only be estimated by Monte Carlo simulation. We assume that the context space $\mathcal{X}$ contains a finite number of $q$ possible contexts ${\boldsymbol x}_{1},\cdots,{\boldsymbol x}_{q}$, and there is a unique set of top-$m$ designs in a certain context $\boldsymbol{x}_{\ell}$, $\ell = 1,\cdots,q$, i.e., $y_{{{\left\langle 1 \right\rangle }_{\ell}}}\left(\boldsymbol{x}_{\ell}\right) \ge \cdots \ge y_{{{\left\langle m \right\rangle }_{\ell}}}\left(\boldsymbol{x}_{\ell}\right) > y_{{{\left\langle m+1 \right\rangle }_{\ell}}}\left(\boldsymbol{x}_{\ell}\right) \ge \cdots \ge y _{{{\left\langle k \right\rangle }_{\ell}}}\left(\boldsymbol{x}_{\ell}\right)$, where ${{\left\langle i \right\rangle }_\ell}$, $i=1,\cdots,k$, $\ell=1,\cdots,q$, are indices ranked by performances of context-dependent designs. Our aim is to correctly select the top-$m$ designs for all possible values of context $\boldsymbol{x}$, i.e., identifying $\mathcal{F}_{\ell}^m \mathop  = \limits^\Delta  \left\{ {{{\left\langle 1 \right\rangle }_\ell},{{\left\langle 2 \right\rangle }_\ell}, \cdots ,{{\left\langle m \right\rangle }_\ell}} \right\}$, $\ell=1,\cdots,q$. See Figure \ref{figconill} for an illustration. For example, in medical resource allocation, we aim to determine the top-$m$ allocation decisions (designs) with the shortest waiting time for critical patients in each emergency patient flow (context). The framework includes different values of $m$ and different number of designs in each context. However, for simplicity of the notation, we assume that both $m$ and the number of designs are the same among each context.

Let ${Y_{i \ell}^h}$, $i = 1,\cdots,k$, $\ell = 1,\cdots,q$, $h \le T$, be the $h$-th independent and identically (i.i.d.) simulation replication for design $i$ in context ${\boldsymbol{x}}_\ell$, where $T< \infty$ is the total simulation budget, and we suppress ${\boldsymbol {x}}_\ell $ in the notation of ${Y_{i \ell}^h}$ for simplicity. We assume that for each design-context pair, the simulation replications follow i.i.d. normal sampling distribution, i.e., $Y_{i \ell}^h \sim N\left( {{y_i}\left( {{{\boldsymbol{x}}_\ell}} \right),\sigma _i^2\left( {{{\boldsymbol{x}}_\ell}} \right)} \right)$, and replications among different designs and contexts are independent. In this work, we assume $\sigma _i^2\left( {{{\boldsymbol{x}}_\ell}} \right)$ in the sampling distribution is known, and use sample estimates as plug-in estimators for the true values in practice. The normality assumption is a common assumption in R\&S literature, and for non-normal sampling distributions, the normality assumption can be justified by batching. A Bayesian framework is introduced to learn unknown parameters $y_i\left({\boldsymbol{x}}_{\ell}\right)$, and the prior distribution is assumed to be a conjugate prior $N ( {y_i^{\left( 0 \right)}\left( {{\boldsymbol{x}_\ell}} \right),{( {\sigma_i^{\left( 0 \right)}\left( {{\boldsymbol{x}_\ell}} \right)} )^2}} )$, where the hyper-parameter $( {\sigma_i^{\left( 0 \right)}\left( {{\boldsymbol{x}_\ell}} \right)} )^2$ quantifies the prior uncertainty. The posterior distribution of $y_i\left({\boldsymbol{x}}_{\ell}\right)$, $i=1,\cdots,k$, $\ell=1,\cdots,q$, after allocating $t$, $t = 1,\cdots, T$, simulation replications is $N (y _i^{\left( t \right)}\left( {{\boldsymbol{x}_\ell}} \right), ( {\sigma _i^{\left( t \right)}\left( {{\boldsymbol{x}_\ell}} \right)} )^2 )$, where
\begin{equation}\label{posteriorvar}
{( {\sigma _i^{\left( t \right)}\left( {{\boldsymbol{x}_\ell}} \right)} )^2} = {\left( {\frac{1}{{{( {\sigma _i^{\left( 0 \right)}\left( {{\boldsymbol{x}_\ell}} \right)} )^2}}} + \frac{{{t_{i\ell}}}}{{\sigma _i^2\left( {{\boldsymbol{x}_\ell}} \right)}}} \right)^{ - 1}}~,
\end{equation}
\begin{equation}\label{posteriormean}
y _i^{\left( t \right)}\left( {{\boldsymbol{x}_\ell}} \right) = {( {\sigma _i^{\left( t \right)}\left( {{\boldsymbol{x}_\ell}} \right)} )^2}\left( {\frac{{y _i^{\left( 0 \right)}\left( {{\boldsymbol{x}_\ell}} \right)}}{{{( {\sigma _i^{\left( 0 \right)}\left( {{\boldsymbol{x}_\ell}} \right)}) ^2}}} + \frac{{{t_{i\ell}}{\bar Y}_{i \ell}}}{{\sigma _i^2\left( {{\boldsymbol{x}_\ell}} \right)}}} \right)~,
\end{equation}
where ${\bar Y}_{i \ell} = \sum\nolimits_{h = 1}^{{t_{i\ell}}} {{{Y_{i \ell}^h} / {{t_{i\ell}}}}}$ is the sample mean, and ${t_{i\ell}}$ is the number of allocated samples to design $i$ under context $\ell$. If ${\sigma _i^{\left( 0 \right)}\left( {{\boldsymbol{x}_\ell}} \right)} \to \infty$, then $y _i^{\left( t \right)}\left( {{\boldsymbol{x}_\ell}} \right) = {\bar Y}_{i \ell}$. In this case, the prior is uninformative, and the posterior distribution only includes sample information without any subjective hyper-parameters. Notice that the posterior mean $y _i^{\left( t \right)}\left( {{\boldsymbol{x}_\ell}} \right)$ is a weighted average between $y _i^{\left( 0 \right)}\left( {{\boldsymbol{x}_\ell}} \right)$ and ${\bar Y}_{i \ell}$, and the posterior variance ${( {\sigma _i^{\left( t \right)}\left( {{\boldsymbol{x}_\ell}} \right)} )^2}$ is non-increasing as $t$ increases, which decreases the uncertainty on $y_i\left({\boldsymbol{x}}_{\ell}\right)$ gradually.

A sampling policy intelligently allocates a fixed simulation budget among design-context pairs, and the estimated top-$m$ context-dependent designs are selected based on collected samples. The unknown parameters $y_i\left({\boldsymbol{x}}_{\ell}\right)$ need to be estimated sequentially from samples in implementation, and thus, the sampling decision essentially becomes a policy adaptive to the information filtration generated by allocated samples. A dynamic sampling policy is a sequence of mappings $\mathcal{A}_{T}\left(\cdot\right) \mathop  = \limits^\Delta \left(A_{1}\left(\cdot\right),\cdots,A_{T}\left(\cdot\right)\right)$, where $A_{t}\left(\mathcal{E}_{t-1}\right) \in \left\{\left(i,\ell \right):\;i\in\left\{1,\cdots,k\right\},\;\ell\in\left\{1,\cdots,q\right\}\right\}$ allocates the $t$-th simulation replication to estimate the performance of design $i$ in context ${\boldsymbol x}_\ell$ based on information filtration $\mathcal{E}_{t-1}$, where ${\mathcal{E}_0}$ contains all hyper-parameters of prior distributions of all design-context pairs. Denote ${t_{i \ell}}\mathop  = \limits^\Delta  \sum\nolimits_{h = 1}^t {A_{i\ell}^{h}} \left( {{\mathcal{E}_{h - 1}}} \right)$ as the corresponding number of allocated samples throughout $t$ steps, where $A_{i\ell}^h\left( {{\mathcal{E}_{h - 1}}} \right)\mathop  = \limits^\Delta  \mathds{1} \left( {{A_h}\left( {{\mathcal{E}_{h - 1}}} \right) = \left( {i,\ell} \right)} \right)$, and $\mathds{1}\left(\cdot\right)$ is an indicator function that equals to 1 when the event in the bracket is true and equals to 0 otherwise. We suppress the dependency of $t_{i\ell}$ on $\mathcal{E}_{t-1}$ for simplicity of the notation. Then the information filtration becomes ${\mathcal{E}_t}\mathop  = \limits^\Delta \{ {{\mathcal{E}_0},Y_{11}^{\left( t \right)}, \cdots ,Y_{1q}^{\left( t \right)}, \cdots ,Y_{k1}^{\left( t \right)}, \cdots ,Y_{kq}^{\left( t \right)}} \}$, $t = 1, \cdots, T$, where $Y_{i \ell}^{\left( t \right)}\mathop  = \limits^\Delta ({Y_{i\ell}^1,Y_{i\ell}^2, \cdots ,Y_{i\ell}^{{t_{i\ell}}}})$ includes sample observations for the design-context pair $\left(i,\ell\right)$. Under conjugate priors, ${\mathcal{E}_t}$ can be completely determined by the posterior hyper-parameters $y_i^{\left( t \right)}\left( {{\boldsymbol{x}_\ell}} \right)$ and $( {\sigma _i^{\left( t \right)}\left( {{\boldsymbol{x}_\ell}} \right)} )^2$. Conditional on $A_{t+1} \left(\mathcal{E}_{t} \right) = (\widehat{i}, \widehat{\ell} )$ and ${\mathcal{E} _{t+1}} =  {\mathcal{E}_t} \cup Y_{\widehat{i}\widehat{\ell}}^{{t_{\widehat{i}\widehat{\ell}}} + 1}$, $\widehat{i} \in \left\{1,\cdots,k\right\}$, $\widehat{\ell} \in \left\{1,\cdots,q\right\}$, if $( { i}, \ell ) \ne (\widehat{i},\widehat{\ell})$, $( {\sigma _{i}^{\left( {t+1} \right)}}\left( {{\boldsymbol{x}_\ell}} \right) )^2$ and ${y_{i}^{\left( {t+1} \right)}}\left( {{\boldsymbol{x}_\ell}} \right)$ remain the same as $( {\sigma _{i}^{\left( t \right)}}\left( {{\boldsymbol{x}_\ell}} \right) )^2$ and ${y_{i}^{\left( {t} \right)}}\left( {{\boldsymbol{x}_\ell}} \right)$, respectively; otherwise,
$$\left( {\sigma _{i}^{\left( {t+1} \right)}}\left( {{\boldsymbol{x}_\ell}} \right) \right)^2 =
\left( {\frac{1}{ ( {\sigma _{i}^{( 0 )}}\left( {{\boldsymbol{x}_\ell}} \right) )^2 } + \frac{{{t_{i\ell}+1}}}{\sigma_{i}^2\left( {{\boldsymbol{x}_\ell}} \right)}} \right)^{-1}$$
$${y_{i}^{\left( {t+1} \right)}}\left( {{\boldsymbol{x}_\ell}} \right) =
{( {\sigma _i^{\left( t+1 \right)}\left( {{\boldsymbol{x}_\ell}} \right)} )^2}\left( {\frac{{y_{i}^{\left( {t} \right)}}\left( {{\boldsymbol{x}_\ell}} \right)}{ ( {\sigma _{i}^{(t)}}\left( {{\boldsymbol{x}_\ell}} \right) )^2 } + \frac{Y_{i\ell}^{t_{i\ell}+1}}{\sigma_{i}^2\left( {{\boldsymbol{x}_\ell}} \right)}} \right)~.$$

We fix a selection policy that in each context, the designs with the top-$m$ largest posterior means are selected after exhausting all $T$ simulation replications. Denote the set of selected top-$m$ designs in context $\boldsymbol{x}_{\ell}$ as $\widehat{\mathcal{F}}_{\ell T}^m \mathop  = \limits^\Delta  \left\{ {{{\left\langle 1 \right\rangle }_{\ell T}},{{\left\langle 2 \right\rangle }_{\ell T}}, \cdots ,{{\left\langle m \right\rangle }_{\ell T}}} \right\}$, $\ell =1,\cdots,q$, where ${{\left\langle i \right\rangle }_{\ell T}}$, $i=1,\cdots,k$ are indices ranked by posterior means. For a certain context $\boldsymbol{x}_{\ell}$, a correct selection occurs when $\widehat{\mathcal{F}}_{\ell t}^m = \mathcal{F}_{\ell}^m$, $\ell \in \left\{1,\cdots,q\right\}$, and the quality of the selection can be measured by
$$\begin{aligned}
& {\rm PCS}\left(\boldsymbol{x}_{\ell}\right) = {\rm Pr}\left\{ \left. \widehat{\mathcal{F}}_{\ell T}^{m}=\mathcal{F}_{\ell}^{m} \right| \mathcal{E}_{T}\right\} \\
= & {\rm Pr}\left\{ \left. \bigcap\nolimits_{i=1}^{m} \bigcap\nolimits_{j=m+1}^{k}\left(y_{\left\langle i \right\rangle_{\ell T}}^{\left(T\right)}\left( {{\boldsymbol{x}_\ell}} \right) > y_{\left\langle j \right\rangle_{\ell T}}^{\left(T\right)}\left( {{\boldsymbol{x}_\ell}} \right)\right) \right| \mathcal{E}_{T}\right\}~.
\end{aligned}$$

The PCS in our problem is also context-dependent. Considering robustness of decision-making, we adopt the ${\rm PCS}_{W}$ to measure the quality of the selection over the entire context space $\mathcal{X}$, i.e., ${\rm PCS}_{W} =\min_{\ell=1,\cdots, q} {\rm PCS}\left(\boldsymbol{x}_{\ell}\right)$. Neither $\rm{PCS}\left({\boldsymbol{x}_{\ell}}\right)$, $\ell=1,\cdots,q$ nor $\rm{PCS}_W$ has an analytical form. We aim to propose a dynamic sampling policy $\mathcal{A}_{T}\left(\cdot\right)$ to maximize the ${\rm PCS}_{W}$ for selecting top-$m$ designs for all contexts.

\section{Efficient Dynamic Sampling Policy}\label{sec3}

The sequential sampling decision can be formulated as a stochastic control problem \cite{peng2016dynamic,peng2018ranking}. The expected payoff for a dynamic sampling policy $\mathcal{A}_{T}\left(\cdot\right)$ in the stochastic control problem can be defined recursively by
$V_{T}\left(\mathcal{E}_{T} ; \mathcal{A}_{T}\left(\cdot\right)\right) \triangleq \mathop {\min}\limits_{\ell=1, \cdots, q} {\rm{Pr}} \left\{ \left. \bigcap_{i=1}^{m} \bigcap_{j=m+1}^{k} \left( y_{{\left\langle i \right\rangle}_{\ell T}}^{\left(T\right)}\left( {{\boldsymbol{x}_\ell}} \right)>y_{{\left\langle j \right\rangle}_{\ell T}}^{\left(T\right)}\left( {{\boldsymbol{x}_\ell}} \right) \right)\right| \mathcal{E}_{T}\right\}$, which is the weighted average of the $\rm{PCS}_W$ under different parameters values, i.e., integrated posterior $\rm{PCS}_W$ ($\rm{IPCS}_W$), and for $0 \le t < T$, $h \in \left\{1,\cdots,k\right\}$, $\ell \in \left\{1,\cdots,q\right\}$, $V_{t}\left(\mathcal{E}_{t} ; \mathcal{A}_{T}\left(\cdot\right)\right) \triangleq \left. \mathbb{E}\left[\left.V_{t+1}\left(\mathcal{E}_{t} \cup \{Y_{h \ell}^{t_{h \ell}+1}\} ; \mathcal{A}_{T}\left(\cdot\right)\right) \right| \mathcal{E}_{t}\right]\right|_{(h, \ell)=A_{t+1}\left(\mathcal{E}_{t}\right)}$.

An optimal dynamic sampling policy $\mathcal{A}_{T}^{*}\left(  \cdot  \right)$ can be defined as the solution of the stochastic control problem, i.e., $\mathcal{A}_{T}^{*}\left(  \cdot  \right)\mathop = \limits^\Delta  \arg \mathop {\max }\nolimits_{{\mathcal{A}_T}\left(  \cdot  \right)} V_0\left(\mathcal{E}_{0}; \mathcal{A}_{T}\left(\cdot\right)\right)$. Using backward induction to solve the stochastic control problem suffers from curse-of-dimensionality. To derive a dynamic sampling policy with an analytical form, we adopt approximate dynamic programming (ADP) scheme, which finds a single feature of the value function one-step look ahead for allocating the next simulation replication, and use certainty equivalence to approximate the value function $V_{T}\left(\mathcal{E}_{T} ; \mathcal{A}_{T}\left(\cdot\right)\right)$.

Suppose any step $t$ could be the last step. Following a VFA scheme in \cite{zhang2021asymptotically}, for a certain context ${\boldsymbol{x}}_{\ell}$, $\ell \in \left\{1,\cdots,q\right\}$, ${\rm Pr}\left\{ \left. \bigcap_{i=1}^{m} \bigcap_{j=m+1}^{k}\left(y_{\left\langle i \right\rangle_{\ell t}}^{\left(T\right)}\left( {{\boldsymbol{x}_\ell}} \right) > y_{\left\langle j \right\rangle_{\ell t}}^{\left(T\right)}\left( {{\boldsymbol{x}_\ell}} \right)\right) \right| \mathcal{E}_{t}\right\}$ is an integral of the density of a $[m\times\left(k-m\right)]$-dimensional standard normal distribution over an area covered by some hyperplanes. The integral over a maximum inscribed sphere in the covered area can well represent the integral due to exponential decay of the normal density. Then an approximation for $V_{t}\left(\mathcal{E}_{t} ; \mathcal{A}_{t}\left(\cdot\right)\right)$ is given by ${\widetilde V_{t}}\left( {{\mathcal{E} _{t}}} \right) \mathop  = \limits^\Delta \mathop {\min }\limits_{\ell = 1, \cdots ,q} {\rm{APCS}}_{\ell}\left( {{\mathcal{E} _{t}}} \right)$, where the radius of the maximum inscribed sphere
$${\rm{APCS}}_{\ell}\left( {{\mathcal{E} _{t}}} \right)  \mathop  = \limits^\Delta \mathop {\min }\limits_{\scriptstyle i = 1, \cdots ,m\atop \scriptstyle j = m + 1,  \cdots ,k} {\frac{\left({y_{{{\left\langle i \right\rangle }_{\ell t}}}^{\left( {t} \right)}\left( {{\boldsymbol{x}_\ell}} \right) - y_{{{\left\langle j \right\rangle }_{\ell t}}}^{\left( {t} \right)}\left( {{\boldsymbol{x}_\ell}} \right)}\right)^2}{ {{\left( {\sigma _{{{\left\langle i \right\rangle }_{\ell t}}}^{\left( {t} \right)}}\left( {{\boldsymbol{x}_\ell}} \right) \right)^2} + {\left( {\sigma _{{{\left\langle j \right\rangle }_{\ell t}}}^{\left( {t} \right)}}\left( {{\boldsymbol{x}_\ell}} \right) \right)^2}} }}~,$$
is considered as a feature of ${\rm {PCS}}\left({\boldsymbol{x}_{\ell}}\right)$. After allocating a simulation replication to design $h$ in context ${\boldsymbol x}_r$, $h\in\left\{1,\cdots,k\right\}$, $r \in \left\{1,\cdots,q\right\}$, a certainty equivalency \cite{bertsekas1995dynamic} is then applied to the VFA one-step look ahead at step $t$: $\mathbb{E} [ { {{{\widetilde V}_{t + 1}}\left( {{\mathcal{E} _t} \cup \{Y_{h r}^{t_{h r}+1}}\} \right)} {|\mathcal{E} _t}} ] \approx \widetilde{V}_{t+1}\left(\mathcal{E}_{t} \cup \mathbb{E}\left[\left. Y_{h r}^{t_{h r}+1} \right| \mathcal{E}_{t}\right]\right)
 = \mathop {\min }\limits_{\ell = 1, \cdots ,q} {\rm{APCS}}_{\ell}\left( {{\mathcal{E} _{t+1}^{E}}} \right)$, where ${\mathcal{E} _{t+1}^{E}} \mathop  = \limits^\Delta \mathcal{E}_{t} \cup \mathbb{E}\left[\left. Y_{h r}^{t_{h r}+1} \right| \mathcal{E}_{t}\right]$, and
\begin{equation}\label{APCS}
\begin{aligned}
& {\rm{APCS}}_{\ell}\left( {\mathcal{E} _{t+1}^{E}} \right)  \mathop   = \limits^\Delta  \\
& \mathop {\min }\limits_{\scriptstyle i = 1, \cdots ,m\atop \scriptstyle j = m + 1,  \cdots ,k} {\frac{\left({y_{{{\left\langle i \right\rangle }_{\ell t}}}^{\left( {t} \right)}\left( {{\boldsymbol{x}_\ell}} \right) - y_{{{\left\langle j \right\rangle }_{\ell t}}}^{\left( {t} \right)}\left( {{\boldsymbol{x}_\ell}} \right)}\right)^2}{ {{\left( {\sigma _{{{\left\langle i \right\rangle }_{\ell (t+1)}}}^{\left( {t+1} \right)}}\left( {{\boldsymbol{x}_\ell}} \right) \right)^2} + { \left( {\sigma _{{{\left\langle j \right\rangle }_{\ell (t+1)}}}^{\left( {t+1} \right)}}\left( {{\boldsymbol{x}_\ell}} \right) \right)^2}} }}~,
\end{aligned}
\end{equation}
where (\ref{APCS}) holds because the additional replication $Y_{h r}^{t_{h r}+1}$ takes a value of its posterior mean $\mathbb{E}\left[\left.Y_{h r}^{t_{h r}+1} \right| \mathcal{E}_{t}\right]$, which yields that all posterior means of $y_{\widetilde i}\left({\boldsymbol x }_{\ell}\right)$, ${\widetilde i} = 1,\cdots,k$ remain unchanged, and only the posterior variance of $y_h\left({\boldsymbol x}_r\right)$ is updated. Notice that allocating a simulation replication to design $h$ in context ${\boldsymbol x}_{r}$ can reduce the posterior variance of $y_h\left(\boldsymbol{x}_r\right)$, and $\widetilde{V}_{t+1}\left(\mathcal{E}_{t} \cup \mathbb{E}\left[\left. Y_{h r}^{t_{h r}+1} \right| \mathcal{E}_{t}\right]\right)$ captures the worst-case performance over all contexts. Let $\widehat{V}_{t}\left(\mathcal{E}_{t} ; \left(h,r\right)\right) \mathop  = \limits^\Delta \widetilde{V}_{t+1}\left(\mathcal{E}_{t} \cup \mathbb{E}\left[\left. Y_{h r}^{t_{h r}+1} \right| \mathcal{E}_{t}\right]\right)$, and then an asymptotically optimal sampling policy for selecting top-$m$ context-dependent designs (AOAmc) that maximizes a VFA one-step look ahead is given by for $h=1,\cdots,k$, $r=1,\cdots,q$, and $t = 0,\cdots,T-1$,
\begin{equation}\label{AOAmc}
{A_{t + 1}}\left( {\mathcal{E} _t} \right) \in
\arg \mathop {\max }\limits_{\left( {h,r} \right)} {{\widehat{V}}_t} \left( {{\mathcal{E}_t};\left( {h,r} \right)} \right)~.
\end{equation}

\begin{coro}\label{easyimple}
The sampling rule (\ref{AOAmc}) can be rewritten as ${A_{t + 1}}\left( {\mathcal{E} _t} \right) = (\widehat{i}^{\left(t + 1\right)}, \widehat{\ell}^{\left(t + 1\right)})$, where $\widehat{i}^{\left(t + 1\right)}$ and $\widehat{\ell}^{\left(t + 1\right)}$ are the design and the context that receive the $\left(t + 1\right)$-th simulation replication, respectively,
\begin{equation}\label{simp1}
{\widehat {\ell}^{\left( t + 1\right)}} \in \arg \mathop {\min }\nolimits_{\ell = 1, \cdots ,q} {APCS_\ell}\left( {{\mathcal{E}_t}} \right)~,
\end{equation}
and let ${H^{\left( t + 1 \right)}}$ be the set containing the indices that achieve ${APCS_{\widehat {\ell}^{\left( t \right)}}}\left( {{\mathcal{E}_t}} \right)$, i.e., for $i = 1, \cdots ,m$, $j = m + 1, \cdots ,k$,
$$\begin{aligned}
& {H^{\left( t + 1 \right)}} =  \\
& \arg \mathop {\min }\limits_{\left\{i,j\right\}} \frac{{{{\left( {y_{{{\left\langle i \right\rangle }_{{{{\widehat \ell }^{\left( t + 1 \right)}}} t}}}^{\left( t \right)}\left( {{{\boldsymbol{x}}_{{{\widehat \ell }^{\left( t + 1\right)}}}}} \right) - y_{{{\left\langle j \right\rangle }_{{{{\widehat \ell }^{\left( t + 1 \right)}}} t}}}^{\left( t \right)}\left( {{{\boldsymbol{x}}_{{{\widehat \ell }^{\left( t + 1 \right)}}}}} \right)} \right)}^2}}}{{{{\left( {\sigma _{{{\left\langle i \right\rangle }_{{{{\widehat \ell }^{\left( t + 1\right)}}} t}}}^{\left( t \right)}\left( {{{\boldsymbol{x}}_{{{\widehat \ell }^{\left( t + 1 \right)}}}}} \right)} \right)}^2} + {{\left( {\sigma _{{{\left\langle j \right\rangle }_{{{{\widehat \ell }^{\left( t + 1\right)}}} t}}}^{\left( t \right)}\left( {{{\boldsymbol{x}}_{{{\widehat \ell }^{\left( t + 1 \right)}}}}} \right)} \right)}^2}}}~.
\end{aligned}$$
Then, $\forall~{h^{\left( t + 1 \right)}} \in {H^{\left( t + 1\right)}}$,
\begin{equation}\label{simpl2}
{{\widehat i}^{\left( t + 1\right)}} \in \arg \mathop {\max }\limits_{{{{\langle h^{\left(t + 1\right)} \rangle }_{{{\widehat \ell }^{\left( t + 1\right)}}t}}}} {{\widehat V}_t} ( {{\mathcal{E} _t}; ( {{{{\langle h^{\left(t + 1\right)} \rangle }_{{{\widehat \ell }^{\left( t + 1 \right)}}t}}},{{\widehat \ell }^{\left( t + 1 \right)}}} )} )~.
\end{equation}
\end{coro}

\begin{remark}
The proof of Corollary \ref{easyimple} can be found in the appendix. According to Corollary \ref{easyimple}, the allocated context ${\widehat {\ell}^{\left( t \right)}}$ following AOAmc achieves the minimal ${APCS_\ell}\left( {{\mathcal{E}_t}} \right)$ among all contexts. The allocated design ${\widehat i}^{\left( t \right)}$ under the context ${\widehat {\ell}^{\left( t \right)}}$ following AOAmc achieves the maximal VFA one-step look ahead among all designs that achieve ${APCS_{\widehat {\ell}^{\left( t \right)}}}\left( {{\mathcal{E}_t}} \right)$. AOAmc focuses on the designs that are most difficult to compare under the context which has the minimal approximate value of PCS among all contexts. Instead of calculating ${{\widehat{V}}_t} \left( {{\mathcal{E}_t};\left( {h,r} \right)} \right)$ for all design-context pairs to determine an allocated design-context pair, Corollary \ref{easyimple} shows that AOAmc requires a time complexity of $O\left(kq+k\right)$ at each step, which is associated with the total number of design-context pairs and the number of designs under the allocated context.
\end{remark}

AOAmc has an analytical form reflecting a mean-variance trade-off. The proposed AOAmc procedure is shown in Algorithm \ref{algAOAmc}. The initial $n_0$ replications are allocated to each design-context pair for estimating their unknown parameters. Additional replications are sequentially allocated to the design-context pair until all replications are exhausted. AOAmc also applies when the value of $m$ and number of designs vary across contexts; see supplementary experiments in Section A.3 of the online appendix \cite{zhang2022online} for example.

\begin{algorithm}[!h]
\caption{AOAmc}
\label{algAOAmc}
 \begin{algorithmic}[1]
 \renewcommand{\algorithmicrequire}{\textbf{Input:}}
 \renewcommand{\algorithmicensure}{\textbf{Output:}}
 \REQUIRE $k$, $q$, $m$, $n_0$, $T$. \\
 \textbf{INITIALIZE:} $t=0$ and perform $n_0$ simulation replications for each design-context pair. \\
 \textbf{LOOP WHILE} $t < \left(T-k \times q \times n_0\right)$, \textbf{DO:}\\
 \quad \textbf{UPDATE:} Update the information filtration according to (\ref{posteriorvar}) and (\ref{posteriormean}).\\
 \quad \textbf{ALLOCATE:} Determine the allocated design-context pair based on (\ref{simp1}) and (\ref{simpl2}). \\
 \quad \textbf{SIMULATE:} Run additional simulation for the allocated design-context pair. \\
 \quad $t=t+1$. \\
\textbf{END OF LOOP.}
\end{algorithmic}
\end{algorithm}

The next result, as stated in Theorem \ref{thmconsistency}, characterizes that as the simulation budget goes to infinity, the top-$m$ designs for all contexts can be correctly identified by AOAmc.

\begin{thm}\label{thmconsistency}
The proposed sampling policy for selecting top-$m$ context-dependent designs (\ref{AOAmc}) is consistent, i.e.,
$$\mathop {\lim }\limits_{t \to \infty } \widehat {\mathcal{F}}_{ \ell t }^m = {\mathcal{F}_{\ell}^m},\;\;a.s.,\;\ell=1,\cdots,q~.$$
\end{thm}

\begin{remark}
The proof of Theorem \ref{thmconsistency} is relegated to Section A.1 of the online appendix \cite{zhang2022online}. The proof of consistency for the top-$m$ context-dependent selection problem (Theorem \ref{thmconsistency}) differs from the proof for the top-$m$ context-free selection problem (Theorem 1 of \cite{zhang2021asymptotically}) in the following aspects: First, we demonstrate that for a fixed context, all designs are sampled infinitely often almost surely. Second, we establish the generalization to any context by using a proof by contradiction, where the definition of ${PCS}_{W}$ is employed to reach contradictions.
\end{remark}

\section{Asymptotic Optimality}\label{sec4}

In addition to consistency, we also prove the asymptotic optimality of our proposed sampling policy. In this section, we begin by deriving the static asymptotically optimal sampling ratios under general sampling distributions by maximizing the large deviations rate of the ${\rm{PFS}}_{W}$. The asymptotic sampling ratios of AOAmc are then shown to be asymptotically optimal under normal sampling distributions.



Define sampling ratios ${r_{h\ell}}\mathop  = \limits^\Delta  {{{t_{h\ell}}}\mathord{\left/
 {\vphantom {{{t_{h\ell}}} T}} \right.\kern-\nulldelimiterspace} T}$, $\sum\nolimits_{h = 1}^k {\sum\nolimits_{\ell = 1}^q {{r_{h\ell}}} }  = 1$, $r_{h\ell} \ge 0$, and ${\boldsymbol r} = \left(r_{11},\cdots,r_{1q},\cdots,r_{k1},\cdots,r_{kq}\right)$ is a vector of the sampling ratios. Under a static optimization framework, where designs $\left\langle h \right\rangle$, $h = 1,\cdots,k$ in each context $\boldsymbol{x}_\ell$ are assumed to be known exactly, the ${\rm PFS}_{W}$ is calculated as ${\rm PFS}_{W} =  \max \limits_{\ell=1, \cdots, q} {\rm Pr}\left\{\bigcup_{i=1}^{m} \bigcup_{j=m+1}^{k} \left(\bar{Y}_{{\left\langle i \right\rangle}\ell}\left( {{r_{\left\langle i \right\rangle \ell}}T} \right) \leq \bar{Y}_{{\left\langle j \right\rangle}{\ell}}\left( {{r_{\left\langle j \right\rangle \ell}}T} \right)\right)\right\}$.

Denote ${\Lambda_{h \ell}}\left( \lambda  \right) \mathop  = \limits^\Delta \log \mathbb{E} ( {{\rm exp}({\lambda {\bar Y_{h \ell}}\left( {{r_{h \ell}}T} \right)})} )$ as the cumulant generating function of sample mean $\bar Y_{h \ell}\left( {{r_{h \ell}}T} \right)$, $h \in \left\{1,\cdots,k\right\}$, $\ell \in \left\{1,\cdots,q\right\}$. The effective domain of ${\Lambda_{h \ell}}\left(  \cdot  \right)$ is ${\mathcal{D}_{{\Lambda_{h \ell}}}} \mathop  = \limits^\Delta \left\{ {\lambda  \in \mathbb{R}:{\Lambda_{h \ell}}\left( \lambda  \right) < \infty } \right\}$. Define ${\mathcal{F}_{h \ell}} \mathop  = \limits^\Delta  \left\{ {{{d{\Lambda_{h \ell}}\left( \lambda  \right)} \mathord{\left/{\vphantom {{d{\Lambda_{h \ell}}\left( \lambda  \right)} {d\lambda }}} \right.\kern-\nulldelimiterspace} {d\lambda }}:\lambda  \in {\mathcal{D}_{\Lambda_i}^{o}}} \right\}$, where $\mathbb{A}^o$ denotes the interior for set $\mathbb{A}$. We make the following technical assumptions in our analysis.

\begin{assumption}\label{ass1}
${\Lambda_{h \ell}}\left( \lambda  \right) = \mathop {\lim }\limits_{T \to \infty } \frac{1}{T}{\Lambda_{h \ell}}\left( {T\lambda } \right)$ is well defined as an extended real number for all $\lambda$, and $0 \in {\mathcal{D}_{\Lambda_{h \ell}^o}}$, $h=1,\cdots,k$, $\ell=1,\cdots,q$.
\end{assumption}

\begin{assumption}\label{ass2}
${\Lambda_{h \ell}}\left( \lambda \right)$ is strictly convex, continuous on ${\mathcal{D}_{\Lambda_h}^o}$ and steep, i.e., $\mathop {\lim }\limits_{T \to \infty } \left| {\Lambda_{h \ell}^{\prime} \left( {{\lambda _T}} \right)} \right| = \infty$, $h=1,\cdots,k$, $\ell=1,\cdots,q$, where $\{\lambda_T\} \in \mathcal{D}_{\Lambda_h}$ is a sequence converging to a boundary point of $\mathcal{D}_{\Lambda_h}^o$.
\end{assumption}

\begin{assumption}\label{ass3}
The interval $\left[ {{y _ {{\left\langle k \right\rangle}_{\ell}}\left(\boldsymbol{x}_{\ell}\right)},{y _ {{\left\langle 1 \right\rangle}_{\ell}}\left(\boldsymbol{x}_{\ell}\right)}} \right] \subset \bigcap\nolimits_{i = 1}^k {\mathcal{F}_{i \ell}^o}$, $\ell=1,\cdots,q$.
\end{assumption}

By G$\rm{\ddot a}$rtner-Ellis theorem \cite{zeitouni2010large}, Assumptions \ref{ass1} and \ref{ass2} indicate that the sample mean ${{\bar Y}_{h \ell}}$ satisfies a large deviations principle with a rate function $\Lambda_{h \ell}^*\left(  x  \right) = \mathop {\sup }_{\lambda  \in \mathbb{R}} \left\{ {\lambda x - {\Lambda_{h \ell}}\left( \lambda  \right)} \right\}$. Assumption \ref{ass2} indicates that $\Lambda_{h \ell}^*\left(x\right)$ are strictly convex and continuous for $x\in{\mathcal{F}_{h \ell}^o}$. Assumption~\ref{ass3} ensures that the sample mean of each design-context pair can take any value in $[ {{y _ {{\left\langle k \right\rangle}_{\ell}}\left(\boldsymbol{x}_{\ell}\right)},{y _ {{\left\langle 1 \right\rangle}_{\ell}}\left(\boldsymbol{x}_{\ell}\right)}} ]$, and $\Pr \{ {{\bar Y}_{\left\langle j \right\rangle \ell}}\left( {{r_{\left\langle j \right\rangle \ell}}T} \right) \ge {{\bar Y}_{\left\langle i \right\rangle \ell}\left( {{r_{\left\langle i \right\rangle \ell}}T} \right)} \} >0$, $i = 1,\cdots ,m$ and $j = m+1,\cdots ,k$. Then we can use the large deviations paradigm to analyze the convergence rate of the ${{\rm{PFS}}_W}$.

The ${{\rm{PFS}}_W}$ is lower bounded by
$$lb \mathop  = \limits^\Delta   \mathop {\max }\limits_{\ell = 1, \cdots ,q} \mathop {\max }\limits_{\scriptstyle i = 1, \cdots ,m\atop
\scriptstyle j = m + 1, \cdots ,k} \Pr \left\{ {{{\bar Y}_{\left\langle i \right\rangle \ell}\left( {{r_{\left\langle i \right\rangle \ell}}T} \right)} \le {{\bar Y}_{\left\langle j \right\rangle \ell}}\left( {{r_{\left\langle j \right\rangle \ell}}T} \right)} \right\}~,$$
and is upper bounded by ${m \times \left(k-m\right)} \times lb$. Although this upper bound could be larger than 1, it shows that the convergence rate of $lb$ itself characterizes the convergence rate of ${{\rm{PFS}}_W}$. Following the G{\"a}rtner-Ellis Theorem \cite{zeitouni2010large}, for ${\bar Y}_{\left\langle j \right\rangle \ell}\left( {{r_{\left\langle j \right\rangle \ell}}T} \right) < {\bar Y}_{\left\langle i \right\rangle \ell}\left( {{r_{\left\langle i \right\rangle \ell}}T} \right)$, there exists a rate function $\mathcal{R}_{ij}^{\ell}\left( {{r_{\left\langle i \right\rangle \ell}},{r_{\left\langle j \right\rangle \ell}}} \right)$ such that $\mathcal{R}_{ij}^{\ell}\left( {{r_{\left\langle i \right\rangle \ell}},{r_{\left\langle j \right\rangle \ell}}} \right) = -  \mathop {\lim }\limits_{T \to \infty } \frac{1}{T}\log \Pr \left\{ {{{\bar Y}_{\left\langle j \right\rangle \ell}}\left( {{r_{\left\langle j \right\rangle \ell}}T} \right) \ge {{\bar Y}_{\left\langle i \right\rangle \ell}\left( {{r_{\left\langle i \right\rangle \ell}}T} \right)}} \right\}$. \cite{zhang2021asymptotically} further shows that $\mathcal{R}_{ij}^{\ell}\left( {{r_{\left\langle i \right\rangle \ell}},{r_{\left\langle j \right\rangle \ell}}} \right) = {r_{\left\langle i \right\rangle \ell}}\Lambda_{{\left\langle i \right\rangle} \ell}^*\left(  x\left( {{r_{\left\langle i \right\rangle \ell}},{r_{\left\langle j \right\rangle \ell}}} \right)  \right) + {r_{\left\langle j \right\rangle \ell}}\Lambda_{{\left\langle j \right\rangle} \ell}^*\left(  x\left( {{r_{\left\langle i \right\rangle \ell}},{r_{\left\langle j \right\rangle \ell}}} \right) \right)$, where $x\left( {{r_{\left\langle i \right\rangle \ell}},{r_{\left\langle j \right\rangle \ell}}} \right)$ is a unique solution satisfying
${r_{\left\langle i \right\rangle \ell}}{\left. {\frac{{d\Lambda_{{\left\langle i \right\rangle} \ell}^*\left( x \right)}}{{dx}}} \right|_{x = x( {{r_{\left\langle  i \right\rangle \ell}},{r_{\left\langle  j \right\rangle \ell}}} )}} + {r_{\left\langle  j \right\rangle \ell}}{\left. {\frac{{d\Lambda_{{\left\langle j \right\rangle} \ell}^*\left( x \right)}}{{dx}}} \right|_{x = x( {{r_{\left\langle  i \right\rangle \ell}},{r_{\left\langle  j \right\rangle \ell}}})}} = 0$. Then the convergence rate of the ${\rm PFS}_{W}$ can be expressed as
$$\begin{aligned}
& \mathop {\lim }\limits_{T \to \infty } \frac{1}{T}\log\left(\mathop {\max }\limits_{\ell} \mathop {\max }\limits_{i,j} \Pr \left\{ {{\bar Y}_{\left\langle j \right\rangle \ell}}\left( {{r_{\left\langle j \right\rangle \ell}}T} \right) \ge {{\bar Y}_{\left\langle i \right\rangle \ell}\left( {{r_{\left\langle i \right\rangle \ell}}T} \right)} \right\}\right) \\
 = & \mathop {\max }\limits_{\ell} \mathop {\max }\limits_{i,j}\mathop {\lim }\limits_{T \to \infty } \frac{1}{T}\log\left(\Pr \left\{ {{\bar Y}_{\left\langle j \right\rangle \ell}}\left( {{r_{\left\langle j \right\rangle \ell}}T} \right) \ge {{\bar Y}_{\left\langle i \right\rangle \ell}\left( {{r_{\left\langle i \right\rangle \ell}}T} \right)} \right\}\right) \\
 = & - \mathop {\min }\limits_{\ell = 1, \cdots,q} \mathop {\min }\limits_{\scriptstyle i = 1, \cdots ,m\atop
\scriptstyle j = m + 1,  \cdots ,k} \mathcal{R}_{ij}^{\ell}\left( {{r_{{\left\langle i \right\rangle \ell}}},{r_{{\left\langle j \right\rangle \ell}}}} \right) \mathop  = \limits^\Delta -\mathcal{R}\left(\boldsymbol{r}\right)~.
\end{aligned}$$

Therefore, ${\rm PFS}_{W}$ for selecting top-$m$ context-dependent designs is shown to converge exponentially with a rate function of $\boldsymbol{r}$, i.e., $- \mathop {\lim }\nolimits_{T \to \infty } \frac{1}{T}\log {\rm PFS}_W = \mathcal{R}\left(\boldsymbol{r}\right)$. Notice that $\mathcal{R}\left(\boldsymbol{r}\right)$ is defined as the slowest rate functions of the most difficult pairwise comparison among all contexts. To derive the asymptotically optimal sampling ratios under a static optimization problem, we consider the following optimization problem:
\begin{alignat}{2}\label{opt0}
\max\quad & \mathcal{R}\left(\boldsymbol{r}\right) \\
\mbox{s.t.}\quad
& \sum\limits_{h = 1}^k {\sum\limits_{\ell = 1}^q {{r_{{\left\langle h \right\rangle} \ell}}} }  = 1, \nonumber\\
& r_{{\left\langle h \right\rangle} \ell} \ge 0,\quad {h=1,\cdots,k},\;{\ell= 1,\cdots q} \nonumber~,
\end{alignat}
where the constraint is equivalent to $\sum\nolimits_{h = 1}^k {\sum\nolimits_{\ell = 1}^q {{t_{{\left\langle h \right\rangle}\ell}}} }  = T$. An equivalent formulation of (\ref{opt0}) can be expressed as
\begin{alignat}{2}\label{opt1}
\max\quad &z \\
\mbox{s.t.}\quad
& \mathcal{R}_{ij}^{\ell}\left( {{r_{{\left\langle i \right\rangle \ell}}},{r_{{\left\langle j \right\rangle \ell}}}} \right) - z \ge 0,\quad {i = 1,\cdots ,m},\nonumber\\
&\quad \quad \quad \quad {j = m + 1, \cdots ,k},\; {\ell = 1,\cdots,q}, \nonumber\\
& \sum\limits_{h = 1}^k {\sum\limits_{\ell = 1}^q {{r_{{\left\langle h \right\rangle}\ell}}} }  = 1,\nonumber\\
& r_{{\left\langle h \right\rangle} \ell} \ge 0,\quad h=1,\cdots,k \nonumber~.
\end{alignat}

$\mathcal{R}_{ij}^{\ell}\left( {\cdot,\cdot} \right)$ is a strictly concave and continuous function of $\left( {{r_{{\left\langle i \right\rangle \ell}}},{r_{{\left\langle j \right\rangle \ell}}}} \right)$. Let ${\widetilde {\mathcal{R}}}_{ij}^{\ell}\left( \boldsymbol{r} \right) = \mathcal{R}_{ij}^{\ell}\left( {{r_{{\left\langle i \right\rangle \ell}}},{r_{{\left\langle j \right\rangle \ell}}}} \right)$, and then ${\widetilde {\mathcal{R}}}_{ij}^{\ell}\left( \boldsymbol{r} \right)$ is a concave function of $\boldsymbol{r}$. The objective function $\mathcal{R}\left(\boldsymbol{r}\right)$ in (\ref{opt0}) is concave for ${\boldsymbol{r}} \ge 0$, since the minimum of concave functions ${\widetilde {\mathcal{R}}}_{ij}^{\ell}\left( \boldsymbol{r} \right)$ is also concave. Therefore, (\ref{opt1}) is a convex optimization problem. There exists a solution $\boldsymbol{r} = \left({1 \mathord{\left/{\vphantom {1 k}} \right. \kern-\nulldelimiterspace} k}, \cdots, {1\mathord{\left/{\vphantom {1 k}} \right.\kern-\nulldelimiterspace} k}\right)$ and $z = 0$ such that the inequality constraints in (\ref{opt1}) strictly hold, and thus the Slater's conditions hold for (\ref{opt1}). Then we can investigate the Karush-Kuhn-Tucker (KKT) conditions for (\ref{opt1}), which are necessary and sufficient optimality conditions.

\begin{lemma}\label{optsol6}
Under Assumptions \ref{ass1}-\ref{ass3}, the optimal solution to (\ref{opt1}) satisfies, for $i,i' = 1,\cdots,m$, $j,j'=m+1,\cdots,k$, $\ell,\ell'=1,\cdots,q$
\begin{equation}\label{generalbal}
\begin{aligned}
& \mathop {\min }\limits_{j = m + 1,  \cdots ,k} {\mathcal{R}}_{ij}^\ell( {r_{{\left\langle i \right\rangle}\ell}^*,r_{{\left\langle j \right\rangle}\ell}^*} )  = \mathop {\min }\limits_{i' = 1, \cdots ,m} {\mathcal{R}}_{i'j'}^\ell( {r_{{\left\langle i^{\prime} \right\rangle}\ell}^*,r_{{\left\langle j^{\prime} \right\rangle}\ell}^*} ) = \\
& \mathop {\min }\limits_{j = m + 1,  \cdots ,k} {\mathcal{R}}_{ij}^{\ell'}( {r_{{\left\langle i \right\rangle}\ell'}^*,r_{{\left\langle j \right\rangle}\ell'}^*} )
 = \mathop {\min }\limits_{i' = 1, \cdots ,m} {\mathcal{R}}_{i'j'}^{\ell'}( {r_{{\left\langle i^{\prime} \right\rangle}\ell'}^*,r_{{\left\langle j^{\prime} \right\rangle}\ell'}^*} ),
\end{aligned}
\end{equation}
where $r_{h\ell}^{*} > 0$, $h = 1,\cdots,k$, $\ell=1,\cdots,q$ are asymptotically optimal sampling ratios.
\end{lemma}

\begin{IEEEproof}
First, we show that $r_{h {\ell}}^*>0$. For ${r_{h {\ell}}^*} = {1 \mathord{\left/{\vphantom {1 k}} \right. \kern-\nulldelimiterspace} k}$, $h=1,\cdots,k$, ${\ell} = 1,\cdots,q$, we have $z > 0$. If there exists $r_{{\left\langle h \right\rangle} {\ell}}^* = 0$, $h \in \left\{1,\cdots,k\right\}$, ${\ell} \in \left\{1,\cdots,q\right\}$, and then we have ${{\mathcal{R}}_{h j}^{\ell}}( {r_{{\left\langle h \right\rangle}{\ell}}^*,r_{{\left\langle  j \right\rangle} {\ell}}^*}) = 0$ or ${{\mathcal{R}}_{i h}^{\ell}}( {r_{{\left\langle  i \right\rangle}{{\ell}}}^*,r_{{\left\langle h \right\rangle}{{\ell}}}^*} ) = 0$, and $z = 0$, $i=1, \cdots,m$, $j=m+1,\cdots,k$, which contradicts to the objective function aiming to maximize $z$. Therefore, the optimal solution to optimization problem (\ref{opt1}) must satisfy $r_{h {\ell}}^* > 0$. With the KKT conditions, there exist multipliers $\gamma$ and $\lambda_{ij}^{\ell} \ge 0$ such that for $i=1,\cdots,m$, $j=m+1,\cdots,k$, $\ell = 1,\cdots,q$,
$$\sum\nolimits_{h = 1}^k {\sum\nolimits_{{\ell} = 1}^q {r_{{\left\langle  h \right\rangle} {\ell}}^* = 1} }~,$$
\begin{equation}\label{K2}
\sum\nolimits_{i = 1}^m {\sum\nolimits_{j = m + 1}^k {\sum\nolimits_{\ell = 1}^q {\lambda _{ij}^{\ell} = 1} } }~,
\end{equation}
\begin{align}\label{K3}
{\sum\nolimits_{j = m + 1}^k {\lambda _{ij}^{\ell}\left. {\frac{{\partial {\mathcal{R}}_{ij}^{\ell}\left( {y,r_{{\left\langle  j \right\rangle}\ell}^*} \right)}}{{\partial y}}} \right|} _{y = r_{{\left\langle  i \right\rangle}\ell}^*}} & = \gamma~,
\end{align}
\begin{align}\label{K4}
\sum\nolimits_{i = 1}^m {\lambda _{ij}^\ell} {\left. {\frac{{\partial {\mathcal{R}}_{ij}^{\ell}\left( {r_{{\left\langle  i \right\rangle}\ell}^*,y} \right)}}{{\partial y}}} \right|_{y = r_{{\left\langle  j \right\rangle}\ell}^*}} & = \gamma~,
\end{align}
\begin{align}\label{K5}
& \lambda _{ij}^\ell\left[ {z - {\mathcal{R}}_{ij}^\ell\left( {r_{{\left\langle  i \right\rangle}\ell}^*,r_{{\left\langle  j \right\rangle}\ell}^*} \right)} \right]  = 0~.
\end{align}

From (\ref{K2}), we have $\mathop {\max}\limits_{i,j,\ell} \lambda _{ij}^{\ell} > 0$, $i=1,\cdots,m$, $j=m+1,\cdots,k$, $\ell=1,\cdots,q$. Since ${\left. {\frac{{\partial {{\mathcal{R}}_{ij}^{\ell}}\left( {y,r_{{\left\langle  j \right\rangle} \ell}^*} \right)}}{{\partial y}}} \right|_{y = r_{{\left\langle  i \right\rangle} \ell}^*}} > 0$ and ${\left. {\frac{{\partial {{\mathcal{R}}_{ij}^{\ell}}\left( {r_{{\left\langle  i \right\rangle} \ell}^*,y} \right)}}{{\partial y}}} \right|_{y = r_{{\left\langle  j \right\rangle} \ell}^*}} > 0$, it follows that $\gamma  > 0$. As a result, with (\ref{K3}) and (\ref{K4}), $\mathop {\max }\limits_{j = m + 1, \cdots ,k}  {{\lambda _{ij}^{\ell}}}  > 0$, $i=1,\cdots,m$, $\ell=1,\cdots,q$ and $\mathop {\max }\limits_{i = 1, \cdots ,m} {{\lambda _{ij}^{\ell}}}  > 0$, $j=m+1,\cdots,k$, $\ell=1,\cdots,q$. Then (\ref{K5}) yields $\mathop {\min }\limits_{j = m + 1, \cdots ,k} {{\mathcal{R}}_{ij}^{\ell}} ( {r_{{\left\langle  i \right\rangle} \ell }^*,r_{{\left\langle  j \right\rangle} \ell}^*} ) = z$, $i=1,\cdots,m$, $\ell=1,\cdots,q$, and $\mathop {\min }\limits_{i = 1, \cdots ,m} {{\mathcal{R}}_{ij}^{\ell}} ( {r_{{\left\langle  i \right\rangle} \ell}^*,r_{{\left\langle  j \right\rangle} \ell}^*} ) = z$, $j=m+1,\cdots,k$, $\ell=1,\cdots,q$. Summarizing the above, the theorem is proved.
\end{IEEEproof}


Equation (\ref{generalbal}) adjusts the number of simulation replications allocated to the top-$m$ and other $\left(k-m\right)$ designs for a certain context and across different contexts. Notice that some Lagrangian multipliers $\lambda _{ij}^{\ell}$ could equal 0 in the KKT conditions due to the summations in (\ref{K3}) and (\ref{K4}), which yields that (\ref{generalbal}) contains different cases by considering the possible number of $\lambda _{ij}^{\ell} = 0$, i.e., the number of inactive inequality constraints $\mathcal{R}_{ij}^{\ell}\left( {{r_{{\left\langle i \right\rangle \ell}}},{r_{{\left\langle j \right\rangle \ell}}}} \right) - z \ge 0$. The number of cases increases as $k$ and $q$ grow. Taking $k = 4$, $q = 2$, and $m = 2$ for an example, (\ref{generalbal}) becomes
\begin{equation}\label{equmulti}
\begin{aligned}
& \mathop {\min } \left\{ {\mathcal{R}}_{13}^{1} \left( \cdot,\cdot \right) , {\mathcal{R}}_{14}^{1} \left( \cdot,\cdot \right) \right\} = \mathop {\min } \left\{ {\mathcal{R}}_{23}^{1} \left( \cdot,\cdot \right) , {\mathcal{R}}_{24}^{1} \left( \cdot,\cdot \right) \right\} \\
= & \mathop {\min } \left\{ {\mathcal{R}}_{13}^{1} \left( \cdot,\cdot \right) , {\mathcal{R}}_{23}^{1} \left( \cdot,\cdot \right) \right\} = \mathop {\min } \left\{ {\mathcal{R}}_{14}^{1} \left( \cdot,\cdot \right) , {\mathcal{R}}_{24}^{1} \left( \cdot,\cdot \right) \right\} \\
= & \mathop {\min } \left\{ {\mathcal{R}}_{13}^{2} \left( \cdot,\cdot \right) , {\mathcal{R}}_{14}^{2} \left( \cdot,\cdot \right) \right\} = \mathop {\min } \left\{ {\mathcal{R}}_{23}^{2} \left( \cdot,\cdot \right) , {\mathcal{R}}_{24}^{2} \left( \cdot,\cdot \right) \right\} \\
= & \mathop {\min } \left\{ {\mathcal{R}}_{13}^{2} \left( \cdot,\cdot \right) , {\mathcal{R}}_{23}^{2} \left( \cdot,\cdot \right) \right\} = \mathop {\min } \left\{ {\mathcal{R}}_{14}^{2} \left( \cdot,\cdot \right) , {\mathcal{R}}_{24}^{2} \left( \cdot,\cdot \right) \right\}~,
\end{aligned}
\end{equation}
which contains 49 cases by considering all possible scenarios in the $\mathop {\min } \left\{\cdot\right\}$ function. This contains a much larger number of cases compared to $q = 1$, leading to a more complex analysis compared to the context-free problem. Notice that (\ref{generalbal}) is derived from the complementary slackness condition (\ref{K5}), which does not fully characterize all KKT conditions. Specifically, it does not ensure that all Lagrangian multipliers $\lambda_{ij}^{\ell} \ge 0$. However, all Lagrangian multipliers can be solved exactly and demonstrated to be positive when $m = 1$ and $m = k-1$ \cite{gao2019selecting,du2022rate}. In contrast, when $m \ne 1$ and $m \ne k-1$, static sampling ratios obtained from certain cases of (\ref{generalbal}) may result in $\lambda_{ij}^{\ell} < 0$, which violates the KKT conditions. This violation indicates that such sampling ratios are asymptotically non-optimal. Therefore, not all possible cases of (\ref{generalbal}) lead to the asymptotically optimal sampling ratios for a certain top-$m$ context-dependent selection problem. In particular, (\ref{generalbal}) serves only as a necessary condition for the asymptotically optimal sampling ratios of the top-$m$ context-dependent selection problem (in particular, $m \ne 1$ and $m \ne k-1$). In Remark \ref{nonuniexa}, an example with rate functions under normal sampling distributions is provided to support these arguments.

We then investigate the conditions satisfied by the asymptotically optimal sampling ratios under normal sampling distributions in Theorem \ref{asopsa3}.


\begin{thm}\label{asopsa3}
Under Assumptions \ref{ass1}-\ref{ass3}, the asymptotically optimal sampling ratios which optimize large deviations rate of the ${\rm{PFS}}_{W}$ for selecting top-$m$ context-dependent designs with normal underlying distributions satisfy, for $i,i' = 1,\cdots,m$, $j,j'=m+1,\cdots,k$, $\ell,\ell'=1,\cdots,q$,
\begin{equation}\label{balance1}
\begin{aligned}
& \mathop {\min }\limits_{j = m + 1, \cdots ,k} \frac{{{\left( {{y _{{{\left\langle i \right\rangle }_\ell}}\left(\boldsymbol{x}_{\ell}\right)} - {y _{{{\left\langle j \right\rangle }_\ell}}\left(\boldsymbol{x}_{\ell}\right)}} \right)^2}}}{{{{\sigma _{{{\left\langle i \right\rangle }_\ell}}^2}\left(\boldsymbol{x}_{\ell}\right) \mathord{\left/
 {\vphantom {{\sigma _{{{\left\langle i \right\rangle }_\ell}}^2} {r_{i\ell}^*}}} \right.
 \kern-\nulldelimiterspace} {r_{{\left\langle i \right\rangle }\ell}^*}} + {{\sigma _{{{\left\langle j \right\rangle }_\ell}}^2}\left(\boldsymbol{x}_{\ell}\right) \mathord{\left/
 {\vphantom {{\sigma _{{{\left\langle j \right\rangle }_\ell}}^2} {r_{j\ell}^*}}} \right.
 \kern-\nulldelimiterspace} {r_{{\left\langle j \right\rangle }\ell}^*}}}} = \\
& \mathop {\min }\limits_{i' = 1, \cdots ,m} \frac{{{{\left( {{y _{{{\left\langle {i'} \right\rangle }_\ell}}\left(\boldsymbol{x}_{\ell}\right)} - {y _{{{\left\langle {j'} \right\rangle }_\ell}}\left(\boldsymbol{x}_{\ell}\right)}} \right)}^2}}}{{{{\sigma _{{{\left\langle {i'} \right\rangle }_\ell}}^2}\left(\boldsymbol{x}_{\ell}\right) \mathord{\left/
 {\vphantom {{\sigma _{{{\left\langle {i'} \right\rangle }_\ell}}^2} {r_{i'\ell}^*}}} \right.
 \kern-\nulldelimiterspace} {r_{{\left\langle i^{\prime} \right\rangle }\ell}^*}} + {{\sigma _{{{\left\langle j^{\prime} \right\rangle }_\ell}}^2}\left(\boldsymbol{x}_{\ell}\right) \mathord{\left/
 {\vphantom {{\sigma _{{{\left\langle j \right\rangle }_\ell}}^2} {r_{{\left\langle j^{\prime} \right\rangle }\ell}^*}}} \right.
 \kern-\nulldelimiterspace} {r_{{\left\langle j^{\prime} \right\rangle }\ell}^*}}}} = \\
& \mathop {\min }\limits_{j = m + 1, \cdots ,k} \frac{{{{\left( {{y _{{{\left\langle i \right\rangle }_{\ell'}}}\left(\boldsymbol{x}_{\ell'}\right)} - {y _{{{\left\langle j \right\rangle }_{\ell'}}}\left(\boldsymbol{x}_{\ell'}\right)}} \right)}^2}}}{{{{\sigma _{{{\left\langle i \right\rangle }_{\ell'}}}^2}\left(\boldsymbol{x}_{\ell'}\right) \mathord{\left/
 {\vphantom {{\sigma _{{{\left\langle i \right\rangle }_{\ell'}}}^2} {r_{i\ell'}^*}}} \right.
 \kern-\nulldelimiterspace} {r_{{\left\langle i \right\rangle }\ell'}^*}} + {{\sigma _{{{\left\langle j \right\rangle }_{\ell'}}}^2}\left(\boldsymbol{x}_{\ell'}\right) \mathord{\left/
 {\vphantom {{\sigma _{{{\left\langle j \right\rangle }_{\ell'}}}^2} {r_{j\ell'}^*}}} \right.
 \kern-\nulldelimiterspace} {r_{{\left\langle j \right\rangle }\ell'}^*}}}} = \\
 & \mathop {\min }\limits_{i' = 1, \cdots ,m} \frac{{{{\left( {{y _{{{\left\langle {i'} \right\rangle }_{\ell'}}}\left(\boldsymbol{x}_{\ell'}\right)} - {y _{{{\left\langle {j'} \right\rangle }_{\ell'}}}\left(\boldsymbol{x}_{\ell'}\right)}} \right)}^2}}}{{{{\sigma _{{{\left\langle {i'} \right\rangle }_{\ell'}}}^2}\left(\boldsymbol{x}_{\ell'}\right) \mathord{\left/
 {\vphantom {{\sigma _{{{\left\langle {i'} \right\rangle }_{\ell'}}}^2} {r_{{\left\langle {i^{\prime}} \right\rangle }\ell'}^*}}} \right.
 \kern-\nulldelimiterspace} {r_{{\left\langle {i^{\prime}} \right\rangle }\ell'}^*}} + {{\sigma _{{{\left\langle j^{\prime} \right\rangle }_{\ell'}}}^2}\left(\boldsymbol{x}_{\ell'}\right) \mathord{\left/
 {\vphantom {{\sigma _{{{\left\langle j \right\rangle }_{\ell'}}}^2} {r_{{\left\langle {j^{\prime}} \right\rangle }\ell'}^*}}} \right.\kern-\nulldelimiterspace} {r_{{\left\langle {j^{\prime}} \right\rangle }\ell'}^*}}}}~,
\end{aligned}
\end{equation}
\begin{equation}\label{balance2}
\sum_{i=1}^{m}\frac{\left(r_{{\left\langle i \right\rangle } \ell}^{*}\right)^{2}}{\sigma_{\langle i\rangle_{\ell}}^{2}\left(\boldsymbol{x}_{\ell}\right)}=\sum_{j=m+1}^{k} \frac{\left(r_{{\left\langle j \right\rangle } \ell}^{*}\right)^{2}}{\sigma_{\langle j\rangle_{\ell}}^{2}\left(\boldsymbol{x}_{\ell}\right)},\quad \ell=1, \cdots, q~.
\end{equation}
\end{thm}

\begin{IEEEproof}
Following \cite{zeitouni2010large}, for the normal underlying distribution, we have $\Lambda_{{\left\langle i \right\rangle} \ell}^*\left( x \right) = {{{ ( {x - {y_{{{\left\langle i \right\rangle }_\ell}}}\left( {\boldsymbol{x}_{\ell}} \right)} )^2}} / { ( {2\sigma _{{{\left\langle i \right\rangle }_\ell}}^2\left( {\boldsymbol{x}_{\ell}} \right)} )}}$, and then it can be derived that
$${\mathcal{R}}_{ij}^\ell \left( {r_{{\left\langle i \right\rangle}\ell},r_{{\left\langle j \right\rangle}\ell}} \right) = \frac{{{\left( {{y _{{{\left\langle i \right\rangle }_\ell}}\left(\boldsymbol{x}_{\ell}\right)} - {y _{{{\left\langle j \right\rangle }_\ell}}\left(\boldsymbol{x}_{\ell}\right)}} \right)^2}}}{2\left({{{\sigma _{{{\left\langle i \right\rangle }_\ell}}^2}\left(\boldsymbol{x}_{\ell}\right) \mathord{\left/ {\vphantom {{\sigma _{{{\left\langle i \right\rangle }_\ell}}^2} {r_{i\ell}}}} \right.\kern-\nulldelimiterspace} {r_{{\left\langle i \right\rangle }\ell}}} + {{\sigma _{{{\left\langle j \right\rangle}_\ell}}^2}\left(\boldsymbol{x}_{\ell}\right) \mathord{\left/
 {\vphantom {{\sigma _{{{\left\langle j \right\rangle }_\ell}}^2} {r_{j\ell}}}} \right.\kern-\nulldelimiterspace} {r_{{\left\langle j \right\rangle }\ell}}}}\right)}~.$$

Substitute ${\mathcal{R}}_{ij}^\ell \left( {r_{{\left\langle i \right\rangle}\ell},r_{{\left\langle j \right\rangle}\ell}} \right)$, $i=1,\cdots,m$, $j=m+1,\cdots,k$, $\ell=1,\cdots,q$ into (\ref{generalbal}), and we have (\ref{balance1}) holds. Then for each context $\ell \in \left\{1,\cdots,q\right\}$, (\ref{balance2}) holds for all cases in (\ref{balance1}) following a similar proof of the Theorem 3 in \cite{zhang2021asymptotically}.
\end{IEEEproof}

If $q =1$, (\ref{balance1}) and  (\ref{balance2}) correspond to the asymptotic optimality conditions in \cite{zhang2021asymptotically} for selecting top-$m$ designs under a single context. (\ref{balance2}) establishes a certain balance between the simulation budget allocated to the top-$m$ designs and other $\left(k-m\right)$ designs for each context $\boldsymbol{x}_{\ell}$, $\ell = 1,\cdots,q$. Although (\ref{balance1}) and  (\ref{balance2}) are derived under the normality assumption, (\ref{generalbal}) holds for general sampling distributions such as Bernoulli and exponential distributions. However, (\ref{generalbal}) does not fully characterize the asymptotically optimal conditions, and estimating rate functions with empirical quantities may lead to large estimation errors. These computational issues hinder developing a sampling procedure solely based on (\ref{generalbal}).

Let $r_{\ell}^* = \sum\nolimits_{h = 1}^k {r_{h{\ell}}^*} > 0$, $\sum\nolimits_{\ell = 1}^q {r_\ell^*}  = 1$ be the asymptotically optimal sampling ratio for each context $\boldsymbol{x}_{\ell}$, $\ell \in \left\{1,\cdots,q\right\}$, and $z^*$ be the optimal value of the objective function achieved by the asymptotically optimal sampling ratios ${\boldsymbol{r}}^* \mathop  = \limits^\Delta ( {r_{11}^*, \cdots r_{1q}^*, \cdots ,r_{k1}^*, \cdots ,r_{kq}^*} ) > 0$ in the optimization problem (\ref{opt1}). Define $\alpha _{h\ell}^* > 0$, $h=1,\cdots,k$, $\ell \in \left\{1,\cdots,q\right\}$ with $\sum\nolimits_{h = 1}^k {\alpha _{h\ell}^*}  = 1$ as the asymptotically optimal sampling ratios for a corresponding top-$m$ context-free selection problem. Denote $\boldsymbol{\alpha} _\ell^* \mathop  = \limits^\Delta \left( {\alpha _{1\ell}^*,\cdots ,\alpha _{k\ell}^*} \right) > 0$, and let ${z_\ell}\left( {\boldsymbol{\alpha}_\ell^*} \right)$ be the optimal large deviations rate of $PFS$ for selecting top-$m$ context-free designs achieved by $\boldsymbol{\alpha} _\ell^*$. Based on the results established in \cite{zhang2021asymptotically}, under normal sampling distributions, $\boldsymbol{\alpha} _\ell^*$ satisfies
\begin{equation}\label{corsingequ}
\begin{aligned}
& \mathop {\min }\limits_{j = m + 1, \cdots ,k} \frac{{{\left( {{y _{{{\left\langle i \right\rangle }_\ell}}\left(\boldsymbol{x}_{\ell}\right)} - {y _{{{\left\langle j \right\rangle }_\ell}}\left(\boldsymbol{x}_{\ell}\right)}} \right)^2}}}{{{{\sigma _{{{\left\langle i \right\rangle }_\ell}}^2}\left(\boldsymbol{x}_{\ell}\right) \mathord{\left/
 {\vphantom {{\sigma _{{{\left\langle i \right\rangle }_\ell}}^2} {\alpha_{i\ell}^*}}} \right.
 \kern-\nulldelimiterspace} {\alpha_{{\left\langle i \right\rangle }\ell}^*}} + {{\sigma _{{{\left\langle j \right\rangle }_\ell}}^2}\left(\boldsymbol{x}_{\ell}\right) \mathord{\left/
 {\vphantom {{\sigma _{{{\left\langle j \right\rangle }_\ell}}^2} {\alpha_{j\ell}^*}}} \right.
 \kern-\nulldelimiterspace} {\alpha_{{\left\langle j \right\rangle }\ell}^*}}}} \\
= & \mathop {\min }\limits_{i' = 1, \cdots ,m} \frac{{{{\left( {{y _{{{\left\langle {i'} \right\rangle }_\ell}}\left(\boldsymbol{x}_{\ell}\right)} - {y _{{{\left\langle {j'} \right\rangle }_\ell}}\left(\boldsymbol{x}_{\ell}\right)}} \right)}^2}}}{{{{\sigma _{{{\left\langle {i'} \right\rangle }_\ell}}^2}\left(\boldsymbol{x}_{\ell}\right) \mathord{\left/
 {\vphantom {{\sigma _{{{\left\langle {i'} \right\rangle }_\ell}}^2} {\alpha_{i'\ell}^*}}} \right.
 \kern-\nulldelimiterspace} {\alpha_{{\left\langle i^{\prime} \right\rangle }\ell}^*}} + {{\sigma _{{{\left\langle j^{\prime} \right\rangle }_\ell}}^2}\left(\boldsymbol{x}_{\ell}\right) \mathord{\left/
 {\vphantom {{\sigma _{{{\left\langle j \right\rangle }_\ell}}^2} {\alpha_{{\left\langle j^{\prime} \right\rangle }\ell}^*}}} \right.
 \kern-\nulldelimiterspace} {\alpha_{{\left\langle j^{\prime} \right\rangle }\ell}^*}}}}~,
\end{aligned}
\end{equation}
and $z_{\ell}\left({\boldsymbol{\alpha}_\ell^*}\right) = \mathop {\min }\limits_{j = m + 1, \cdots ,k} \frac{{{\left( {{y _{{{\left\langle i \right\rangle }_\ell}}\left(\boldsymbol{x}_{\ell}\right)} - {y _{{{\left\langle j \right\rangle }_\ell}}\left(\boldsymbol{x}_{\ell}\right)}} \right)^2}}}{{{{\sigma _{{{\left\langle i \right\rangle }_\ell}}^2}\left(\boldsymbol{x}_{\ell}\right) \mathord{\left/
 {\vphantom {{\sigma _{{{\left\langle i \right\rangle }_\ell}}^2} {\alpha_{i\ell}^*}}} \right.
 \kern-\nulldelimiterspace} {\alpha_{{\left\langle i \right\rangle }\ell}^*}} + {{\sigma _{{{\left\langle j \right\rangle }_\ell}}^2}\left(\boldsymbol{x}_{\ell}\right) \mathord{\left/
 {\vphantom {{\sigma _{{{\left\langle j \right\rangle }_\ell}}^2} {\alpha_{j\ell}^*}}} \right.
 \kern-\nulldelimiterspace} {\alpha_{{\left\langle j \right\rangle }\ell}^*}}}}$. In addition, for any $\boldsymbol{\widetilde{\alpha}}_\ell^* > 0$ satisfying (\ref{corsingequ}), ${z_\ell}\left( {\boldsymbol{\alpha}_\ell^*} \right) > {z_\ell}\left( {\widetilde{\boldsymbol{\alpha}}_\ell^*} \right)$. The following corollary establishes a connection between the asymptotically optimal sampling ratios of the top-$m$ context-dependent selection problem with $q$ contexts and $q$ independent top-$m$ context-free selection problems under normal sampling distributions.

\begin{coro}\label{saratocon}
Under Assumptions \ref{ass1}-\ref{ass3} and the normality assumption, we have $\alpha _{h \ell}^* = {{r_{h \ell}^*} \mathord{\left/{\vphantom {{r_{h \ell}^*} {r_\ell^*}}} \right.\kern-\nulldelimiterspace} {r_\ell^*}}$, $h = 1,\cdots,k$, $\ell=1,\cdots,q$, where
$$r_\ell^* = \frac{{{1 \mathord{\left/
{\vphantom {1 {{z_\ell}\left( {\boldsymbol{\alpha}_\ell^*} \right)}}} \right.\kern-\nulldelimiterspace} {{z_\ell}\left( {\boldsymbol{\alpha}_\ell^*} \right)}}}}{{\sum\nolimits_{\widetilde \ell = 1}^q {{1 \mathord{\left/{\vphantom {1 {{z_{\widetilde \ell}}\left( {\alpha _{\widetilde \ell}^*} \right)}}} \right.\kern-\nulldelimiterspace} {{z_{\widetilde \ell}}\left( {\alpha _{\widetilde \ell}^*} \right)}}} }},\quad z^* = \frac{1}{{\sum\nolimits_{\widetilde \ell = 1}^q {{1 \mathord{\left/{\vphantom {1 {{z_{\widetilde \ell}}\left( {\alpha _{\widetilde \ell}^*} \right)}}} \right.\kern-\nulldelimiterspace} {{z_{\widetilde \ell}}\left( {\alpha _{\widetilde \ell}^*} \right)}}} }}~.$$
\end{coro}

\begin{remark}
The proof of Corollary \ref{saratocon} can be found in the appendix. According to Corollary \ref{saratocon}, in the top-$m$ context-dependent selection problem with normal underlying distributions, the asymptotically optimal sampling ratio for each context is inversely proportional to the corresponding optimal large deviations rate of the PFS for selecting top-$m$ context-free designs. In addition, the optimal large deviations rate of ${PFS}_{W}$ is associated with the optimal large deviations rates of the $PFS$ for $q$ independent top-$m$ context-free selection problems. Corollary \ref{saratocon} provides a method for calculating the asymptotically optimal sampling ratios of the top-$m$ context-dependent selection problem with $q$ contexts by independently solving the asymptotically optimal sampling ratios of $q$ independent top-$m$ context-free selection problems.
\end{remark}

The following theorem shows that under the normality assumption, the sampling ratios of AOAmc can achieve the asymptotically optimal sampling ratios as the number of simulation budget goes to infinity.

\begin{thm}\label{thmasymptoticoptimality}
Suppose (\ref{balance1}) and  (\ref{balance2}) determine a unique solution. Under Assumptions \ref{ass1}-\ref{ass3} and the normality assumption, as $t \to \infty$, with the proposed AOAmc procedure, the sampling ratio of each design-context pair sequentially achieves the asymptotically optimal sampling ratio which optimizes the large deviations rate of ${PFS}_{W}$, i.e.,
$$\mathop {\lim }\limits_{t \to \infty } r_{\left\langle h \right\rangle \ell}^{\left( t \right)} = r_{\left\langle h \right\rangle \ell}^*,\;a.s.,\quad h=1,\cdots,k,\;\ell=1,\cdots,q~,$$
where $r_{h \ell}^{\left( t \right)} = {{{t_{h \ell}}} \mathord{\left/
 {\vphantom {{{t_{h \ell}}} t}} \right.\kern-\nulldelimiterspace} t}$, $\sum\nolimits_{h = 1}^k {\sum\nolimits_{\ell = 1}^q {r_{h \ell}^{\left( t \right)} = 1} }$, $r_{h \ell}^* > 0$, and ${{\boldsymbol{r}}^*}$ satisfies (\ref{balance1}) and (\ref{balance2}).
\end{thm}

\begin{remark}
The proof of Theorem \ref{thmasymptoticoptimality} is rather technical and lengthy, and it can be found in Section A.1 of the online appendix \cite{zhang2022online}. The key technical differences in the proof of asymptotic optimality between the top-$m$ context-dependent selection problem (Theorem \ref{thmasymptoticoptimality}) and the top-$m$ context-free selection problem (Theorem 4 of \cite{zhang2021asymptotically}) are outlined below. Firstly, in proving the positivity of the convergence ratios and satisfying (\ref{balance1}), we use a proof by contradiction and rely on the definition of ${PCS}_{W}$ to reach contradictions. Secondly, in proving the convergence ratios satisfying (\ref{balance2}), the constructions of sets of implicit functions and systems of equations are more complicated since we need to consider the convergence ratios for all design-context pairs. The assumption of the uniqueness of the solution to (\ref{balance1}) and (\ref{balance2}) is essential for the top-$m$ context-dependent selection problem because not all possible cases of (\ref{balance1}) combined with (\ref{balance2}) lead to asymptotically optimal sampling ratios. This difficulty does not exist when $m = 1$ as shown in \cite{li2020context,du2022rate}, and \cite{gao2019selecting}.
\end{remark}

\begin{remark}\label{nonuniexa}
In the following, we provide an example to demonstrate the potential non-uniqueness of the solution to (\ref{balance1}) and (\ref{balance2}). Take $k = 4$, $q = 2$, and $m = 2$ for an example. The mean performances of all design-context pairs are given as:
$$\begin{aligned}
\boldsymbol{y} & = \left\{ {{y_{{{\left\langle 1 \right\rangle }_1}}}, \cdots ,{y_{{{\left\langle 4 \right\rangle }_1}}},{y_{{{\left\langle 1 \right\rangle }_2}}}, \cdots ,{y_{{{\left\langle 4 \right\rangle }_2}}}} \right\} \\
& = \left\{ {6.7,5.9,3.5,1.9,9.6,8.2,5.4,4.3} \right\}~,
\end{aligned}$$
with variances:
$$\begin{aligned}
{\boldsymbol{\sigma}^2} & = \{ {\sigma _{{{\left\langle 1 \right\rangle }_1}}^2, \cdots ,\sigma _{{{\left\langle 4 \right\rangle }_1}}^2,\sigma _{{{\left\langle 1 \right\rangle }_2}}^2, \cdots ,\sigma _{{{\left\langle 4 \right\rangle }_2}}^2} \} \\
& = \left\{ {3.6,1.1,4.4,4.7,3.7,8.0,6.5,4.6} \right\}~.
\end{aligned}$$
The four solutions to (\ref{balance1}) and (\ref{balance2}) and their corresponding large deviations rate of $PFS_{W}$ are shown in Table \ref{tablerema5}. Solutions 1-4 in Table \ref{tablerema5} correspond to ${\mathcal{R}}_{13}^{1} = {\mathcal{R}}_{14}^{1} = {\mathcal{R}}_{24}^{1}  = {\mathcal{R}}_{13}^{2}  = {\mathcal{R}}_{14}^{2}  = {\mathcal{R}}_{24}^{2} = 0.04748$, ${\mathcal{R}}_{13}^{1} = {\mathcal{R}}_{14}^{1} = {\mathcal{R}}_{24}^{1} = {\mathcal{R}}_{13}^{2} = {\mathcal{R}}_{23}^{2} = {\mathcal{R}}_{24}^{2} = 0.06081$, ${\mathcal{R}}_{13}^{1} = {\mathcal{R}}_{23}^{1} = {\mathcal{R}}_{24}^{1}  = {\mathcal{R}}_{13}^{2}  = {\mathcal{R}}_{14}^{2}  = {\mathcal{R}}_{24}^{2} = 0.05382$, and ${\mathcal{R}}_{13}^{1} = {\mathcal{R}}_{23}^{1} = {\mathcal{R}}_{24}^{1}  = {\mathcal{R}}_{13}^{2}  = {\mathcal{R}}_{23}^{2}  = {\mathcal{R}}_{24}^{2} = 0.07163$, respectively.
Therefore, based on the definition of $\boldsymbol{r}^*$, Solution 4 represents the asymptotically optimal sampling ratios for this example, whereas the other three solutions are asymptotically non-optimal, even though they satisfy (\ref{balance1}) and (\ref{balance2}). In this example, out of the 49 possible cases in (\ref{equmulti}), only 4 cases result in analytical solutions with only one case leading to the asymptotically optimal sampling ratios. The presence of negative values for the corresponding Lagrangian multipliers in Solutions 1-3 further illustrates why these solutions are asymptotically non-optimal. For example, in Solution 1, $\lambda_{14}^{1} = -0.4058$ and $\lambda_{14}^{2} = -0.4564$; in Solution 2, $\lambda_{14}^{1} = -0.4057$; in Solution 3, $\lambda_{14}^{2} = -1.3297$. In order to differentiate the top-$2$ designs from the other 2 designs in each context, the design-context pairs ${{\left\langle 2 \right\rangle }_1}$, ${{\left\langle 3 \right\rangle }_1}$, ${{\left\langle 2 \right\rangle }_2}$, and ${{\left\langle 3 \right\rangle }_2}$ tend to receive more simulation replications. However, Table \ref{tablerema5} reveals that too many simulation replications are allocated to some of the four design-context pairs in the three asymptotically non-optimal solutions. For example, in Solution 1, excessive simulation replications are allocated to all the above four design-context pairs; in Solution 2, excessive simulation replications are allocated to ${{\left\langle 2 \right\rangle }_1}$ and ${{\left\langle 3 \right\rangle }_1}$; in Solution 3, excessive simulation replications are allocated to ${{\left\langle 2 \right\rangle }_2}$ and ${{\left\langle 3 \right\rangle }_2}$. In the top-$m$ context-dependent selection problem, the number of potential asymptotically non-optimal solutions is larger compared to the top-$m$ context-free selection problem (in particular, $m \ne 1$ and $m \ne k-1$). The potential non-uniqueness of the solution to (\ref{balance1}) and (\ref{balance2}) poses a challenge in characterizing the asymptotic optimality of any sampling policy derived for the top-$m$ context-dependent selection problem (in particular, $m \ne 1$ and $m \ne k-1$). However, this challenge does not affect the performance of the proposed AOAmc, since it is not derived based on (\ref{balance1}) and (\ref{balance2}). In Section A.2 of the online appendix \cite{zhang2022online}, we empirically show that based on this example, the proposed AOAmc tends to achieve the asymptotically optimal sampling ratios, rather than the asymptotically non-optimal ones as the simulation budget increases. If the solutions to (\ref{balance1}) and (\ref{balance2}) are not unique, the asymptotic optimality of AOAmc can be numerically checked by verifying that its asymptotic solutions satisfy the KKT conditions in the proof of Lemma \ref{optsol6}.
\begin{table}[!h]
\caption{Four solutions to (\ref{balance1}) and (\ref{balance2}) in the example in Remark \ref{nonuniexa}}
\label{tablerema5}
\centering
\begin{tabular}{ccccc}
\hline
Sampling Ratios  & Solution 1 & Solution 2 & Solution 3 & Solution 4 \\
\hline
$r_{{\left\langle 1 \right\rangle }1}$ & 0.04210 & 0.05393 & 0.05930 & \textbf{0.07892}\\
$r_{{\left\langle 2 \right\rangle }1}$ & 0.09680 & 0.12399 & 0.05762 & \textbf{0.07668}\\
$r_{{\left\langle 3 \right\rangle }1}$ & 0.19700 & 0.25233 & 0.12785 & \textbf{0.17015}\\
$r_{{\left\langle 4 \right\rangle }1}$ & 0.02991 & 0.03831 & 0.03628 & \textbf{0.04829}\\
$r_{{\left\langle 1 \right\rangle }2}$ & 0.02285 & 0.03241 & 0.02590 & \textbf{0.03817}\\
$r_{{\left\langle 2 \right\rangle }2}$ & 0.30417 & 0.23814 & 0.34483 & \textbf{0.28050}\\
$r_{{\left\langle 3 \right\rangle }2}$ & 0.27281 & 0.21059 & 0.30927 & \textbf{0.24805}\\
$r_{{\left\langle 4 \right\rangle }2}$ & 0.03436 & 0.05030 & 0.03895 & \textbf{0.05924}\\
\hline
$z$ & 0.04748 & 0.06081 & 0.05382 & \textbf{0.07163} \\
\hline
\end{tabular}
\end{table}
\end{remark}

\section{Numerical Experiments}\label{sec5}

In this section, we conduct numerical experiments to test the performance of different sampling policies on selection of top-$m$ context-dependent designs. The proposed AOAmc is compared with the equal allocation (EA), optimal computing budget allocation for selecting top-$m$ designs with equal allocation among contexts (E-OCBAm), asymptotically optimal allocation policy for selecting top-$m$ designs with equal allocation among contexts (E-AOAm), top-$m$ identification for linear bandits algorithm (m-LinGapE), and balancing the optimal large deviations rate functions for selecting top-$m$ context-dependent designs (BOLDmc).

Specifically, EA equally allocates simulation replications to estimate the performance of each design-context pair, i.e., roughly ${t_{h\ell}} = T  / \left( {k \times q} \right)$, $h=1,\cdots,k$, $\ell=1,\cdots,q$. The OCBAm procedure \cite{chen2008efficient} is designed to identify the top-$m$ designs in a single context, which requires a separating parameter to differentiate the top-$m$ designs from the other $\left(k-m\right)$ designs. We slightly improve it for the top-$m$ context-dependent selection problem by equally allocating simulation replications among different contexts and running the OCBAm procedure in each context. In the implementation of E-OCBAm, we combine it with a heuristic ``most starving" sequential rule \cite{chen2011stochastic} and use a separating parameter suggested by \cite{zhang2021asymptotically}, which has been shown to improve the performance of the OCBAm procedure. E-AOAm is an extension of the AOAm procedure \cite{zhang2021asymptotically}, which allocates simulation replications equally among different contexts and runs the AOAm procedure for each context. m-LinGapE, proposed by \cite{reda2021top}, first finds a design-context pair $\left(i,\ell \right)$ such that $\mathop {\min }\nolimits_{\ell} \mathop {\max }\nolimits_{i,j} x_\ell^T ({\widehat \theta _{i,t}^{\lambda}} - {\widehat \theta _{j,t}^{\lambda}} )$ and then finds the most competing design-context pair $(j,\widetilde{\ell} )$ against $\left(i,\ell \right)$ such that $\mathop {\min }\nolimits_{\ell,\widetilde{\ell}} \mathop {\max }\nolimits_{i,j} \boldsymbol{x}_\ell^T{\widehat \theta _{i,t}^{\lambda}} - \boldsymbol{x}_{\widetilde{\ell}}^T{\widehat \theta _{j,t}^{\lambda}} + {C_{t\delta }}\sqrt {{{\left( {{\boldsymbol{x}_\ell} - {\boldsymbol{x}_{\widetilde{\ell}}}} \right)}^T}{{\widetilde \sigma }^2}{{\left( {V_t^\lambda } \right)}^{ - 1}}\left( {{\boldsymbol{x}_\ell} - {\boldsymbol{x}_{\widetilde{\ell}}}} \right)}$, where $\widehat \theta _{h,t}^\lambda  = {\left( {V_t^\lambda } \right)^{ - 1}}\sum\nolimits_{\ell' = 1}^q {{{\bar Y}_{h,t}}\left( {{\boldsymbol{x}_{\ell'}}} \right)} {\boldsymbol{x}_{\ell'}}$, $V_t^\lambda  = \lambda {I_d} + \sum\nolimits_{\ell' = 1}^q {{t_{\ell'}}} {\boldsymbol{x}_{\ell'}}\boldsymbol{x}_{\ell'}^T$, ${t_{\ell'}} = \sum\nolimits_{\ell' = 1}^q {{t_{h\ell'}}}$, ${C_{t\delta }} = \sqrt {2\ln \left( {{{\left( {\ln  t  + 1} \right)} / \delta }} \right)} $, and $\widetilde{\sigma}^2$ is the variance of context value. An allocated design-context pair is sequentially determined such that $\mathop {\max }\nolimits_{\ell''} {C_{t\delta }}\sqrt {\boldsymbol{x}_{\ell''}^T{{\widetilde \sigma }^2}{{\left( {V_t^\lambda } \right)}^{ - 1}}{\boldsymbol{x}_{\ell''}}} $, $\ell'' = \ell,\widetilde{\ell}$. Finally, we consider the extension of the CR\&S procedure in \cite{du2022rate} to the top-$m$ context-dependent selection problem. The CR\&S procedure is conceptually similar to the BOLD procedure in \cite{chen2022balancing}, and thus we denote the extended procedure as BOLDmc. BOLDmc allocates simulation replications to sequentially balance each equation of the asymptotic optimality conditions (\ref{balance1}) and (\ref{balance2}). At each step, it first finds design-context pairs $\left( {{{\left\langle i \right\rangle }_{{\ell^*}t}},{{\left\langle j \right\rangle }_{{\ell^*}t}}} \right)$ such that
\begin{equation}\label{boldequ}
\arg \mathop {\min }\limits_{\ell = 1, \cdots ,q} \mathop {\min }\limits_{\scriptstyle i = 1, \cdots ,m \atop \scriptstyle j = m + 1, \cdots ,k} \frac{{{{\left( {{{\bar Y}_{{{\left\langle i \right\rangle }_{\ell t}}\ell}} - {{\bar Y}_{{{\left\langle j \right\rangle }_{\ell t}}\ell}}} \right)}^2}}}{{{{{{\widehat \sigma} _{{{\left\langle i \right\rangle }_{\ell t}}}^2} \mathord{\left/{\vphantom {{{\widehat \sigma} _{{{\left\langle i \right\rangle }_{lt}}}^2} {{t_{{{\left\langle i \right\rangle }_{lt}}l}}}}} \right.\kern-\nulldelimiterspace} {{t_{{{\left\langle i \right\rangle }_{\ell t}}\ell}}}} + {{{\widehat \sigma} _{{{\left\langle j \right\rangle }_{\ell t}}}^2} \mathord{\left/ {\vphantom {{{\widehat \sigma} _{{{\left\langle j \right\rangle }_{\ell t}}}^2} {{t_{{{\left\langle j \right\rangle }_{\ell t}}\ell}}}}} \right.\kern-\nulldelimiterspace} {{t_{{{\left\langle j \right\rangle }_{\ell t}}\ell}}}}}}}~,
\end{equation}
where ${{\widehat \sigma} _{h}^2}$ is the sample variance and then allocates a simulation replication to a design-context pair $\left({{\left\langle i \right\rangle }_{{\ell^*}t}},\ell^*\right)$ if $\sum\limits_{i = 1}^m {\frac{{t_{{{\left\langle i \right\rangle }_{{\ell^*}t}}{\ell^*}}^2}}{{{\widehat \sigma} _{{{\left\langle i \right\rangle }_{{\ell^*}t}}}^2}}}  < \sum\limits_{j = m + 1}^k {\frac{{t_{{{\left\langle j \right\rangle }_{{\ell^*}t}}{\ell^*}}^2}}{{{\widehat \sigma} _{{{\left\langle j \right\rangle }_{{\ell^*}t}}}^2}}}$; otherwise, to $\left({{\left\langle j \right\rangle }_{{\ell^*}t}},\ell^*\right)$.

EA utilizes no sample information. Both E-OCBAm and E-AOAm are derived by solving $q$ independent top-$m$ context-free selection problems, which only utilize the sample information among designs and do not consider the information among contexts. m-LinGapE assumes a linear dependence between the performances of each design and the contexts and utilizes context parameters as auxiliary information. BOLDmc utilizes the information from the sample means and variances of design-context pairs. Following Corollary \ref{easyimple}, when the prior is uninformative, (\ref{boldequ}) in BOLDmc is the same as the right-hand side of (\ref{simp1}), indicating that the allocated context at each step under BOLDmc and AOAmc is the same. The main differences between BOLDmc and AOAmc lie in their approaches to determining the allocated design at each step. The former focuses on asymptotic optimality, whereas the latter pursues both desirable one-step optimlaity and asymptotic optimality. In Section A.2 of the online appendix \cite{zhang2022online}, we empirically test the trend of sampling ratios of BOLDmc as the simulation budget increases based on the example in Remark \ref{nonuniexa}. Numerical results show that, similar to the results of AOAmc in Remark \ref{nonuniexa}, the sampling ratios of BOLDmc also tend to converge to the asymptotically optimal sampling ratios as the simulation budget increases, rather than the asymptotically non-optimal sampling ratios. However, considering that (\ref{balance1}) and (\ref{balance2}) only serve as a necessary condition for the asymptotic optimality of a fixed-budget sampling policy designed for the top-$m$ context-dependent selection problem (in particular, $m \ne 1$ and $m \ne k-1$), potential issues arise regarding the rationality of deriving BOLDmc solely based on the necessary condition. In addition, although BOLDmc tends to achieve the asymptotically optimal sampling ratios in the tested example in Remark \ref{nonuniexa}, which is a simple example since different design-context pairs are easy to differentiate, the existence of non-uniqueness in the solution to (\ref{balance1}) and (\ref{balance2}) may lead to misleading results for BOLDmc in more complex problem settings. Determining whether there are problem settings where the sampling ratios of BOLDmc converge to asymptotically non-optimal sampling ratios is left as an open question. In contrast, AOAmc does not rely on (\ref{balance1}) and (\ref{balance2}) for its derivation, yet it can still be demonstrated to possess desirable asymptotic optimality.

In all numerical experiments, we set the number of initial simulation replications as ${n_{0}} = 10$ for each design-context pair. For m-LinGapE, context parameters $\boldsymbol{x}_{\ell}$ are set as single-dimensional variables generated from $\boldsymbol{x}_{\ell} \sim N\left(5,1^2\right)$. Other parameters involved in m-LinGapE are specified as $\delta = 0.05$ and $\lambda = \widetilde{\sigma} / 20$. In Experiment A.3.3 of the online appendix \cite{zhang2022online}, we empirically demonstrate that allocating one simulation replication per step tends to lead to better performance for both E-OCBAm and BOLDmc when the number of simulation budget is small. In addition, both sampling policies exhibit robustness across the number of incremental simulation replications. Experiments 1-3 (as well as Experiments A.3.1 and A.3.2 of the online appendix \cite{zhang2022online}) include synthetic examples. In order to test the robustness of the performance of AOAmc across different parameter settings, we randomly generate the performance of each design-context pair from a normal prior in each independent macro experiment. The statistical efficiency of each sampling policy is measured by the ${\rm IPCS}_{W}$. Examples 4 and 5 have deterministic performance for each design-context pair, and the statistical efficiency of different sampling policies is measured using the classical ${\rm PCS}_{W}$. All performance metrics are estimated by $100,000$ independent macro experiments. The codes for implementing the numerical experiments can be found at https://github.com/gongbozhang-pku/Context-Dependent-Selection.

\textit{Experiment 1: $10 \times 5$ design-context pairs.} The sampling policies are tested in a synthetic example with 10 competing designs and 5 contexts. We consider selecting top-$3$ designs for each context. In each macro experiment, the performance of each design-context pair is generated from ${y_h}\left( {{\boldsymbol{x}_\ell}} \right) \sim N\left( {0,6^2} \right)$, $h=1,\cdots,10$, $\ell=1,\cdots,5$. The simulation replications are drawn independently from a normal distribution $N\left({y_h}\left( {{\boldsymbol{x}_\ell}} \right),{\sigma_h^2}\left( {{\boldsymbol{x}_\ell}} \right)\right)$, where ${\sigma_h}\left( {{\boldsymbol{x}_\ell}} \right) = 6$. The simulation budget is $T=2500$. The standard deviation in the generating distribution of the true means measures their dispersion. In Experiment 1, the differences in the true means among different competing design-context pairs are of the same order of magnitude as the standard deviation of simulation replications. Statistically speaking, this suggests that the differences in the true means would be relatively large, so Experiment 1 can be categorized as a high-confidence scenario.

\begin{figure}[!h]
\centering
\includegraphics[width=0.41\textwidth]{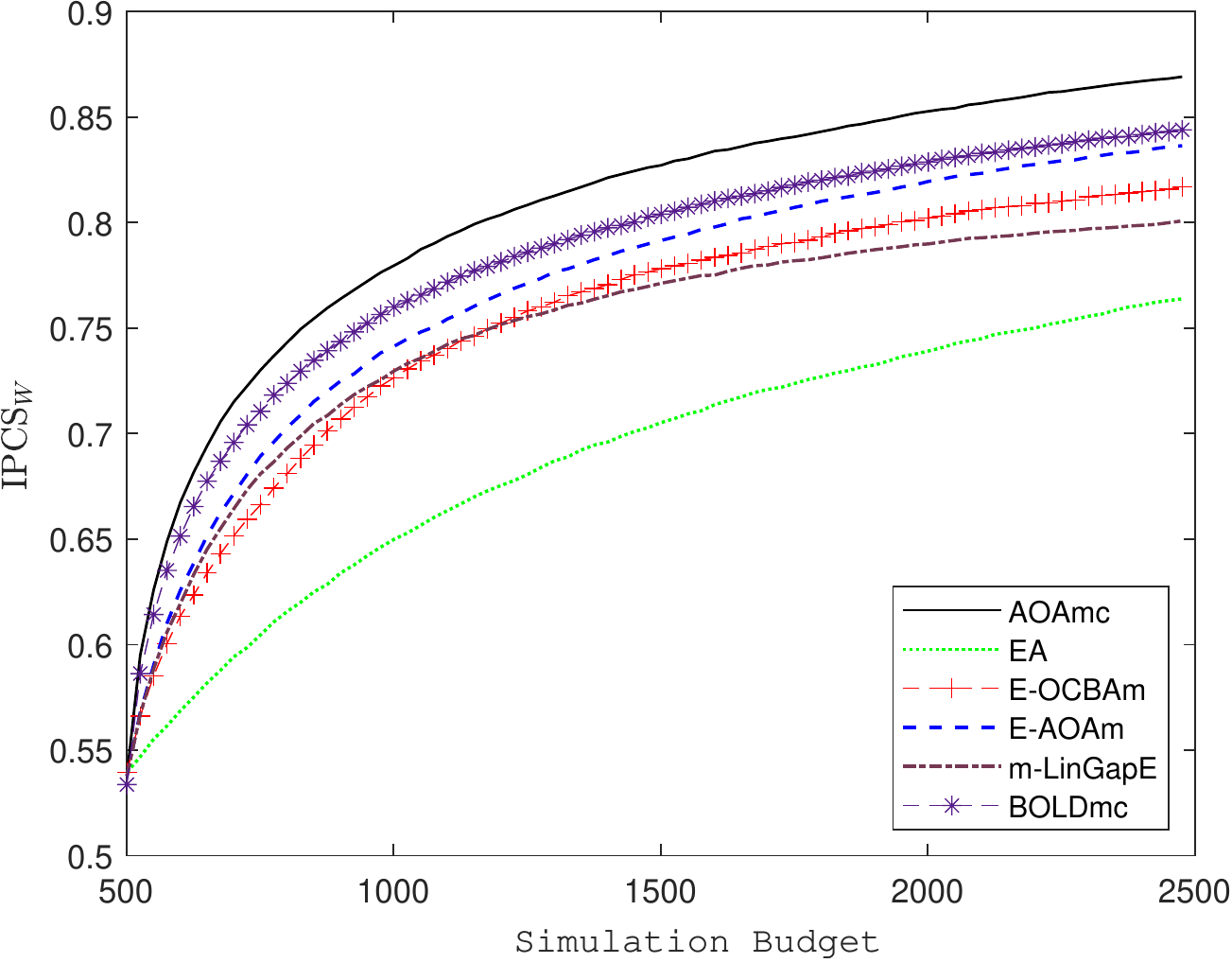}
\caption{${\rm IPCS}_{W}$ of the six sampling policies in Experiment 1.}
\label{fig2}
\end{figure}

From Figure \ref{fig2}, we can observe that EA performs the worst among all sampling policies. m-LinGapE and E-OCBAm have a comparable performance at the beginning, with the latter surpassing the former as the simulation budget increases. E-AOAm has an edge over E-OCBAm and m-LinGapE, and the gap between E-AOAm and BOLDmc narrows as the number of simulation budget grows. BOLDmc outperforms all sampling policies except AOAmc, which performs the best among all sampling policies. The performance enhancement of AOAmc over E-AOAm demonstrates that solving $q$ independent context-free selection problems is inefficient for a context-dependent selection problem, especially when the simulation is expensive. AOAmc consumes 1950 simulation replications to attain ${\rm IPCS}_{W}=0.85$, whereas the other sampling policies require more than 2500 simulation replications to achieve the same ${\rm IPCS}_{W}$ level, i.e., AOAmc reduces the consumption of simulation replications by more than 22.0\%. The performance enhancement of AOAmc could be attributed to the stochastic control framework, which formulates the sequential sampling decision under a finite-sample budget constraint. In addition, AOAmc possesses desirable one-step optimality as well as asymptotic optimality, fully utilizing the sample information among design-context pairs.

To show the effect of the parameter $m$ on different sampling policies, Table \ref{table1} reports the ${\rm IPCS}_{W}$ values of six sampling policies for five different $m$ values: $m = 1,3,5,7,9$, when $T = 2500$. From Table \ref{table1}, we can observe that AOAmc consistently outperforms other sampling policies across all $m$ values, demonstrating its robustness.

\begin{table}[!h]
\caption{${\rm IPCS}_{W}$ of six sampling policies for different $m$ values}
\label{table1}
\centering
\begin{tabular}{cccccccc}
\hline
   $m$ & 1 & 3 & 5 & 7 & 9 \\
\hline
   AOAmc  & 0.7838 & 0.8699  & 0.8335  & 0.8712  & 0.9469 \\
   EA & 0.5320  & 0.7645 & 0.7269  & 0.7632  & 0.8800  \\
   E-OCBAm & 0.6872  & 0.8176  & 0.7834  & 0.8193 & 0.9259 \\
   E-AOAm & 0.6968  &  0.8373  & 0.8039 & 0.8360 & 0.9295 \\
   m-LinGapE & 0.6555 & 0.8011  & 0.7607 & 0.8055 & 0.9183 \\
   BOLDmc & 0.7814 & 0.8449 & 0.8114 & 0.8446 & 0.9454 \\
\hline
\end{tabular}
\end{table}

\textit{Experiment 2: $10 \times 5$ design-context pairs.} The sampling policies are tested in a synthetic example with 10 designs and 5 contexts. The numerical settings are the same as in Experiment 1, except that the performance of each design-context pair is generated from ${y_h}\left( {{\boldsymbol{x}_\ell}} \right) \sim N ( {0,\left(\sqrt {3.6} \right)^2} )$, $h=1,\cdots,10$, $\ell=1,\cdots,5$ in each macro experiment. In Experiment 2, statistically speaking, the differences in the true means among different competing design-context pairs would be relatively small compared to the standard deviation of simulation replications. Therefore, the numerical settings of Experiment 2 can be categorized as a low-confidence scenario.

\begin{figure}[!h]
\centering
\includegraphics[width=0.41\textwidth]{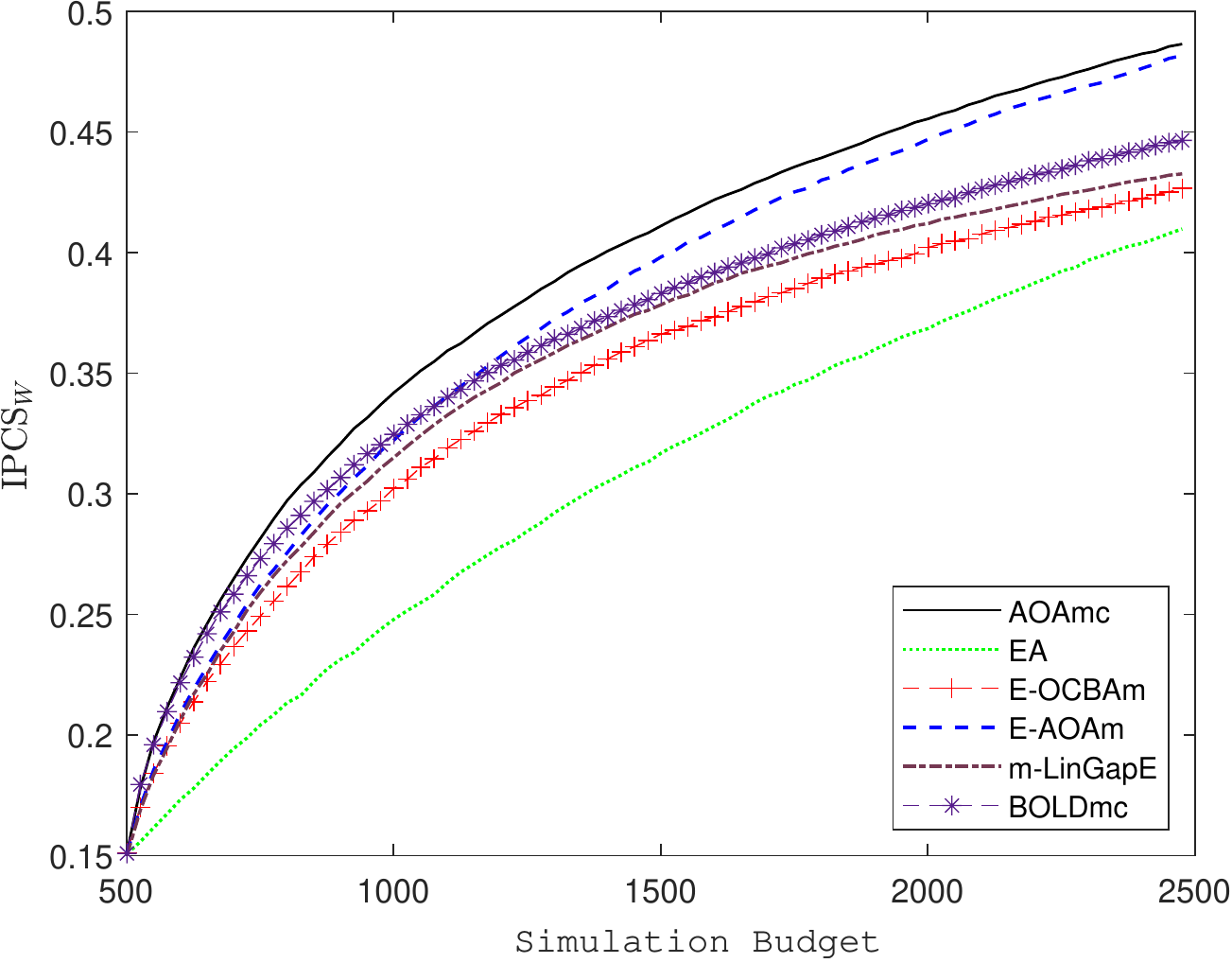}
\caption{${\rm IPCS}_{W}$ of the six sampling policies in Experiment 2.}
\label{low8}
\end{figure}

In Figure \ref{low8}, we can see that the ${\rm IPCS}_{W}$ of EA increases at a slow pace, and the gap between EA and E-OCBAm narrows as the simulation budget increases. E-AOAm, E-OCBAm, and m-LinGapE have a comparable performance at the beginning, whereas m-LinGapE has an edge over E-OCBAm as the simulation budget increases, and E-AOAm performs the best among the three sampling policies as the simulation budget increases. BOLDmc has an edge over E-AOAm at the beginning, with the latter surpassing the former as the number of simulation budget grows. AOAmc has a slight edge over BOLDmc at the beginning, and the gap between them widens as the simulation budget increases. AOAmc is superior to all other sampling policies in comparison. It requires 1930 simulation replications to attain ${\rm IPCS}_{W}=0.45$, whereas the other sampling policies consume more than 2040 simulation replications to attain the same ${\rm IPCS}_{W}$ level, i.e., AOAmc reduces the consumption of simulation replications by more than 5.4\%. The robustness of AOAmc highlights its superiority and potential for practical applications when different design-context pairs are difficult to differentiate.

Comparing Figure \ref{fig2} and Figure \ref{low8}, we can observe that the performances of both BOLDmc and E-OCBAm deteriorate in the low-confidence scenario. This highlights the difference between the asymptotic property and the finite sample behavior of a sequential sampling policy. Specifically, this observation empirically implies that a sequential sampling policy derived based on replacing unknown means and variances with their finite sample estimates in the asymptotic optimality conditions does not guarantee a good finite sample performance.

\textit{Experiment 3: $50 \times 10$ design-context pairs.} The sampling policies are tested in a larger synthetic example with 50 designs and 10 contexts. We consider selecting top-$5$ designs for each context. In each macro experiment, the performance of each design-context pair is generated from ${y_h}\left( {{\boldsymbol{x}_\ell}} \right) \sim N\left( {30,10^2} \right)$, $h=1,\cdots,50$, $\ell=1,\cdots,10$. For design $h$ and context $\ell$, the simulation replications are drawn independently from a normal distribution $N\left({y_h}\left( {{\boldsymbol{x}_\ell}} \right),\sigma _h^2\left( {{\boldsymbol{x}_\ell}} \right)\right)$, where $\sigma _h\left( {{\boldsymbol{x}_\ell}} \right) \sim U\left(4,6\right)$. The simulation budget is $T=10000$. In Experiment 3, the variance of simulation replications is randomly generated for each independent macro experiment, whereas in Experiment 1, it remains constant across all independent macro experiments. The numerical settings of Experiment 3 can be categorized as a high-confidence scenario.

\begin{figure}[!h]
\centering
\includegraphics[width=0.41\textwidth]{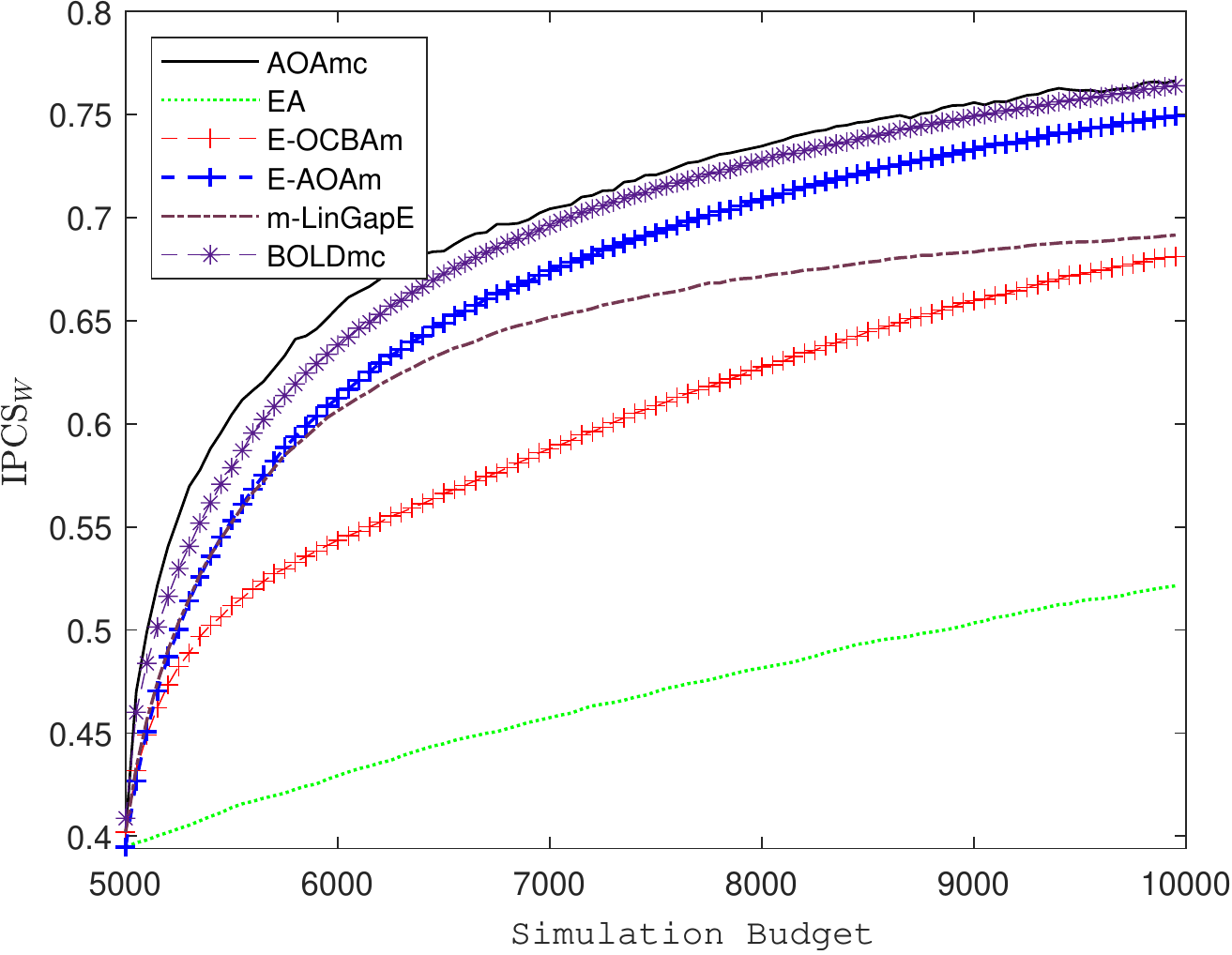}
\caption{${\rm IPCS}_{W}$ of the six sampling policies in Experiment 3.}
\label{fig5}
\end{figure}

In Figure \ref{fig5}, we can see that the ${\rm IPCS}_{W}$ of EA increases at a slow pace when the number of designs and contexts is large. E-OCBAm performs better than EA and catches up with m-LinGapE as the simulation budget increases. E-AOAm and m-LinGapE have a comparable performance at the beginning, and the ${\rm IPCS}_{W}$ of the former increases at a faster pace than the latter as the simulation budget increases. BOLDmc outperforms all the compared sampling policies except for AOAmc, and the gap between BOLDmc and AOAmc narrows as the number of simulation budget grows. AOAmc demonstrates superiority over other sampling policies. It requires 6915 simulation replications to attain ${\rm IPCS}_{W}=0.7$, whereas the other sampling policies consume more than 7095 simulation replications to attain the same ${\rm IPCS}_{W}$ level, i.e., AOAmc reduces the consumption of simulation replications by more than 2.5\%. The robustness of AOAmc highlights its superiority and potential for practical applications in scenarios with a large number of designs and contexts.

\begin{figure}[!h]
\centering
\includegraphics[width=0.33\textwidth]{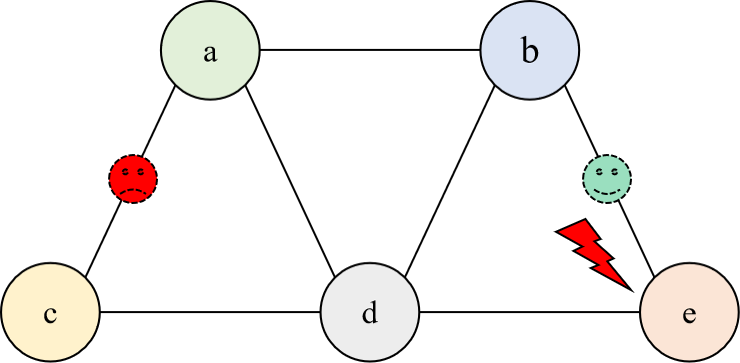}
\caption{Honeypot-enabled network in Experiment 4. If the attacked node is $e$, and the honeypot is placed on edge $be$, the defender gains a capturing payoff; if the honeypot is placed on edge $ac$, the attacker gains a successful attack payoff.}
\label{fig6}
\end{figure}

\textit{Experiment 4: honeypot deception games example}. The devices with wireless connections are vulnerable to malicious hacking attempts. Allocating honeypots over the network is a mechanism of deception to protect against cyber-attacks. Honeypots deceive attackers from reaching their targets and help defenders learn hacking techniques used by attackers. In this example, we use game theory to capture the strategic interactions between a defender and an adversary in a honeypot-enabled network. We consider a network of 5 nodes and 7 edges, as shown in Figure \ref{fig6}, where each node represents a vulnerability associated with a device, and an edge connects two network vulnerabilities. The exact location of the attacker is unknown to the defender, and the attacker does not gather any information about the network before launching attacks. Therefore, the reward of an action pair is unknown in advance. Inputs of the simulation model include design parameters (defensive actions) and context parameters (offensive actions). We aim to determine some defensive actions for designers to protect the network from a known class of attacks. This example has also been considered in \cite{kiekintveld2015game} and \cite{la2016deceptive}. The parameters in the reward matrix are based on \cite{anwar2020honeypot}.

The strategic game is formulated as a two-person zero-sum static game $\left(\mathcal{N},\mathcal{A},\mathcal{R}\right)$, where $\mathcal{N}$ involves a defender and an attacker, $\mathcal{A}$ contains the available game actions, and $\mathcal{R}$ is a reward matrix determined by the action pairs $\left(a_i,a_j\right)$ taken by the two players. The defensive actions involve determining which edge to allocate the honeypot, while the offensive actions involve deciding which node to attack. In addition, the defender has the option to take a no-allocation action, and the attacker has the option to take a no-attack action to avoid potential cost loss, given the uncertainty regarding the opponent's actions.

The honeypot allocation cost is denoted as $P_c$, and the attack cost is denoted as $A_c$. If the defender places a honeypot on the edge connecting the attacked node, they gain a capturing payoff denoted as $Cap$. Otherwise, the attacker gains a successful attack payoff denoted as $Att$. The payoff is weighted by the importance of the attacked node, taking into consideration the varying importance of different nodes in the network. Specifically, the unknown rewards for the defender are represented as $\mathcal{R}_{ij} = y_{ij} + \epsilon_{ij}$, $i=1,\cdots,k$, $j=1,\cdots,q$, where $\epsilon_{ij}$ follows normal distribution, and
$$\begin{aligned}
& {y _{ij}}\left( {{a_i},{a_j}} \right)= \\
& \left\{
\begin{array}{cll}
- {P_c} + {A_c} + Cap*{w_u} &,& {a_i} = {e_{hu}},{a_j} = u, \;h \ne u \\
- {P_c} + {A_c} - Att*{w_u} &,& {a_i} = {e_{h\ell}},{a_j} = u, \; h,\ell \ne u \\
- {P_c} &,& {a_i} = {e_{h\ell}},{a_j} = 0 \\
{A_c} - Att*{w_u} &,& {a_i} = 0,{a_j} = u \\
0 &,& {a_i} = 0,{a_j} = 0 \\
\end{array}
\right.~,
\end{aligned}$$
where $e_{hu}$ is an edge connecting nodes $h$ and $u$, $a_i = 0$ denotes the no-allocation action of the defender, and $a_j = 0$ denotes the no-attack action of the attacker. In this example, there are $k=8$ defensive actions and $q=6$ offensive actions. We consider selecting the top-$3$ defensive actions with the  largest rewards for each offensive action. The parameters are set as follows: $P_c = \left(5,1,6,7,3,4\right) $, $A_c=4$, $Cap = 10$, $Att= 10$, $w=\left(0.2401,0.2669,0.0374,0.2692,0.1864\right)$, and $\epsilon_{ij} \mathop \sim \limits^{i.i.d.} N\left(0,3^2\right)$. The true reward matrix is shown in Table \ref{table2}. The simulation budget is $T=3000$.
\begin{table}[!h]
\caption{The reward matrix with elements $y_{ij}\left(a_i,a_j\right)$ in the honeypot deception games example.}
\label{table2}
\centering
\begin{tabular}{ c  c  c  c  c  c  c }
\hline
$y_{ij}$ & a & b & c & d & e & no-attack \\
\hline
$e_{ab}$ & 1.401 & 1.669 & -1.374 & -3.692 & -2.864 & -5\\
$e_{ac}$ & 5.401 & 0.331 & 3.374 & 0.308 & 1.136 & -1\\
$e_{ad}$ & -5.265 & -4.669 & -2.374 & 0.692 & -3.864 & -6\\
$e_{bd}$ & -5.401 & -0.331 & -3.374 & -0.308 & -4.864 & -7\\
$e_{be}$ & -1.401 & 3.669 & 0.626 & -1.692 & 2.864 & -3 \\
$e_{cd}$ & -0.401 & -0.669 & 2.374 & 4.692 & 0.136 & -2 \\
$e_{de}$ & -2.401 & -2.669 & -0.374 & 2.692 & 1.864 & -4\\
no-allocation & 1.599 & 1.331 & 3.626 & 1.308 & 2.136 & 0 \\
\hline
\end{tabular}
\end{table}

\begin{figure}[!h]
\centering
\includegraphics[width=0.41\textwidth]{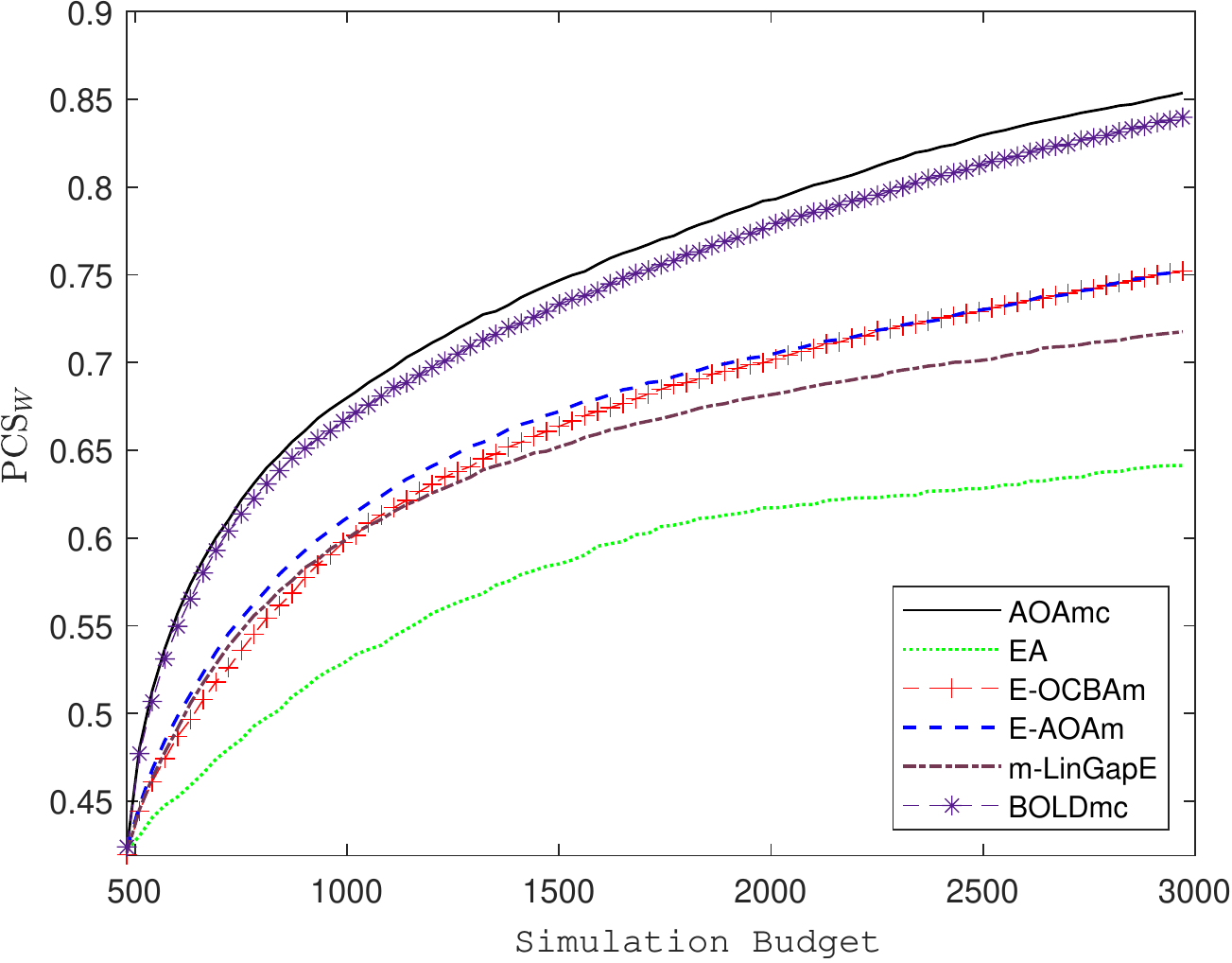}
\caption{${\rm PCS}_{W}$ of the six sampling policies in the honeypot deception games example.}
\label{fig7}
\end{figure}

\begin{figure}[!h]
\centering
\subfigure[AOAmc]{\includegraphics[width=0.158\textwidth]{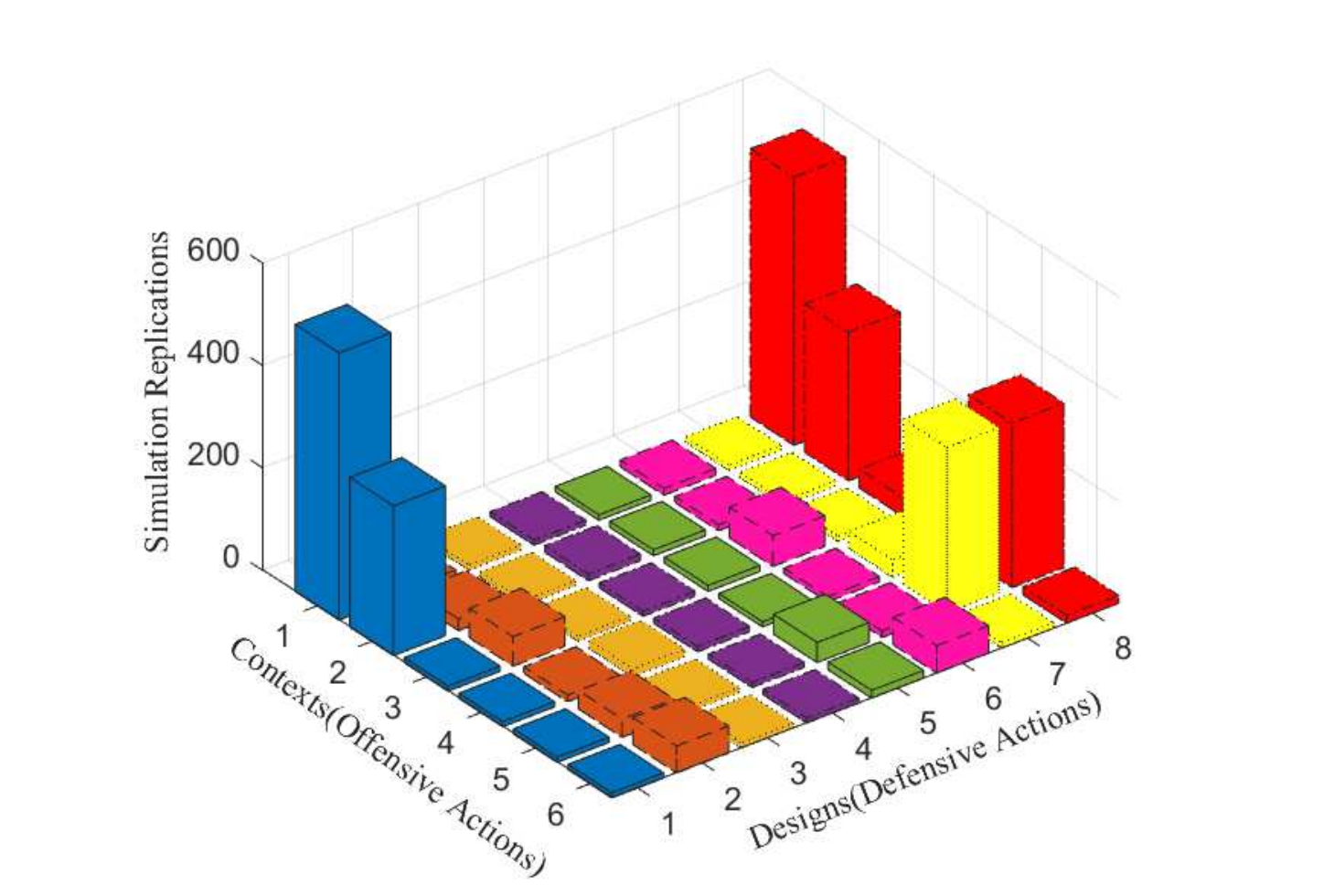}}
\subfigure[EA]{\includegraphics[width=0.158\textwidth]{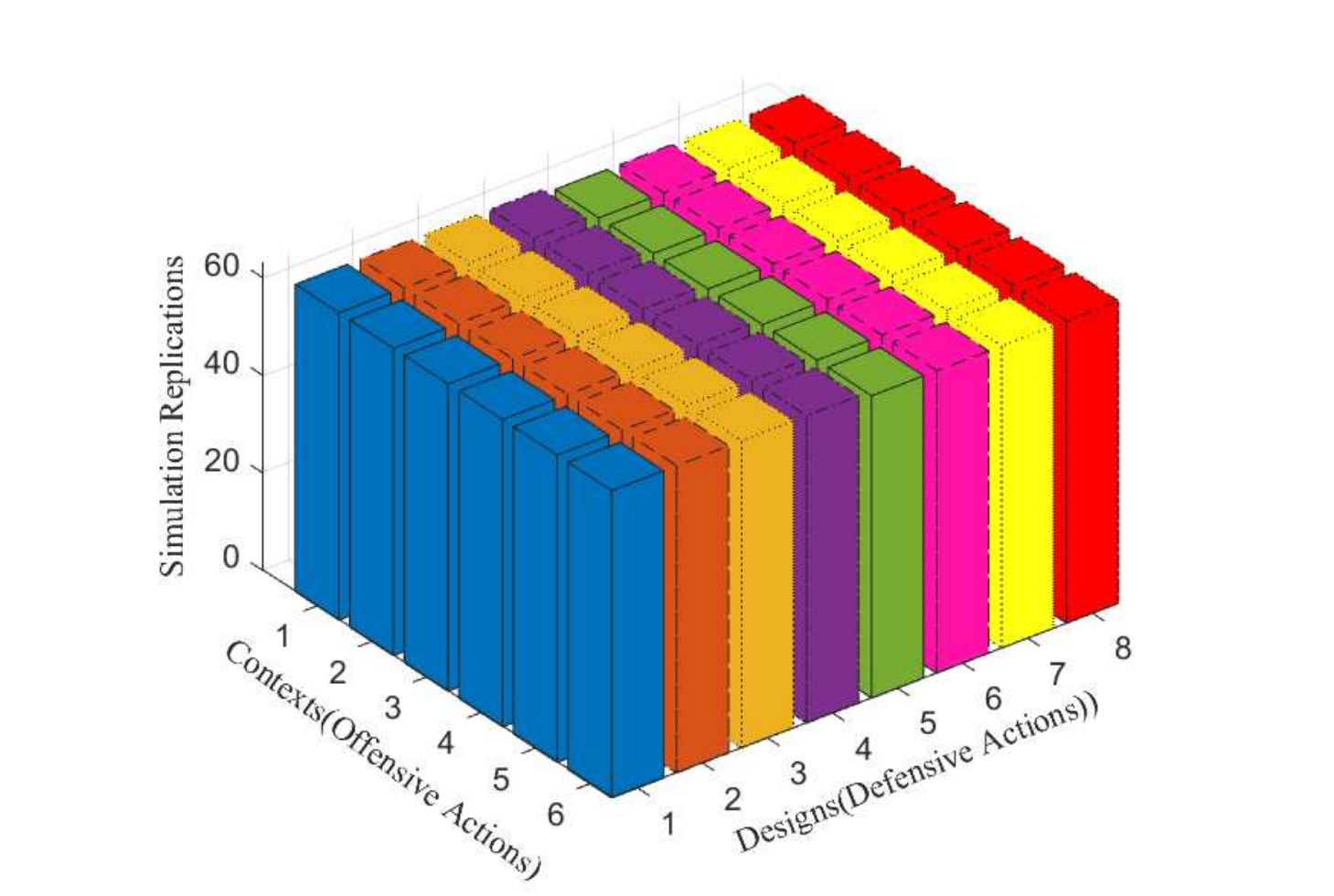}}
\subfigure[E-OCBAm]{\includegraphics[width=0.158\textwidth]{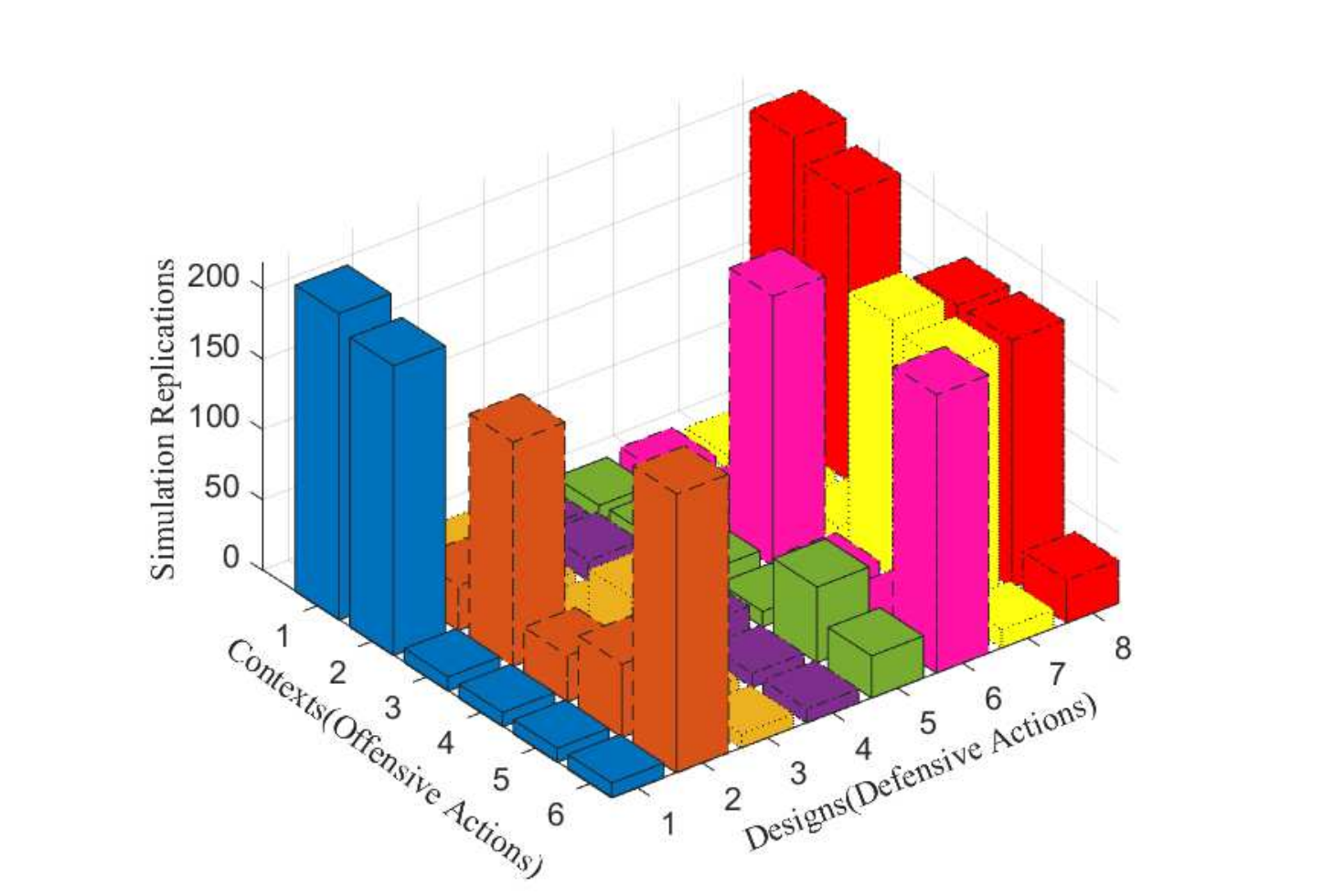}}
\quad
\subfigure[E-AOAm]{\includegraphics[width=0.158\textwidth]{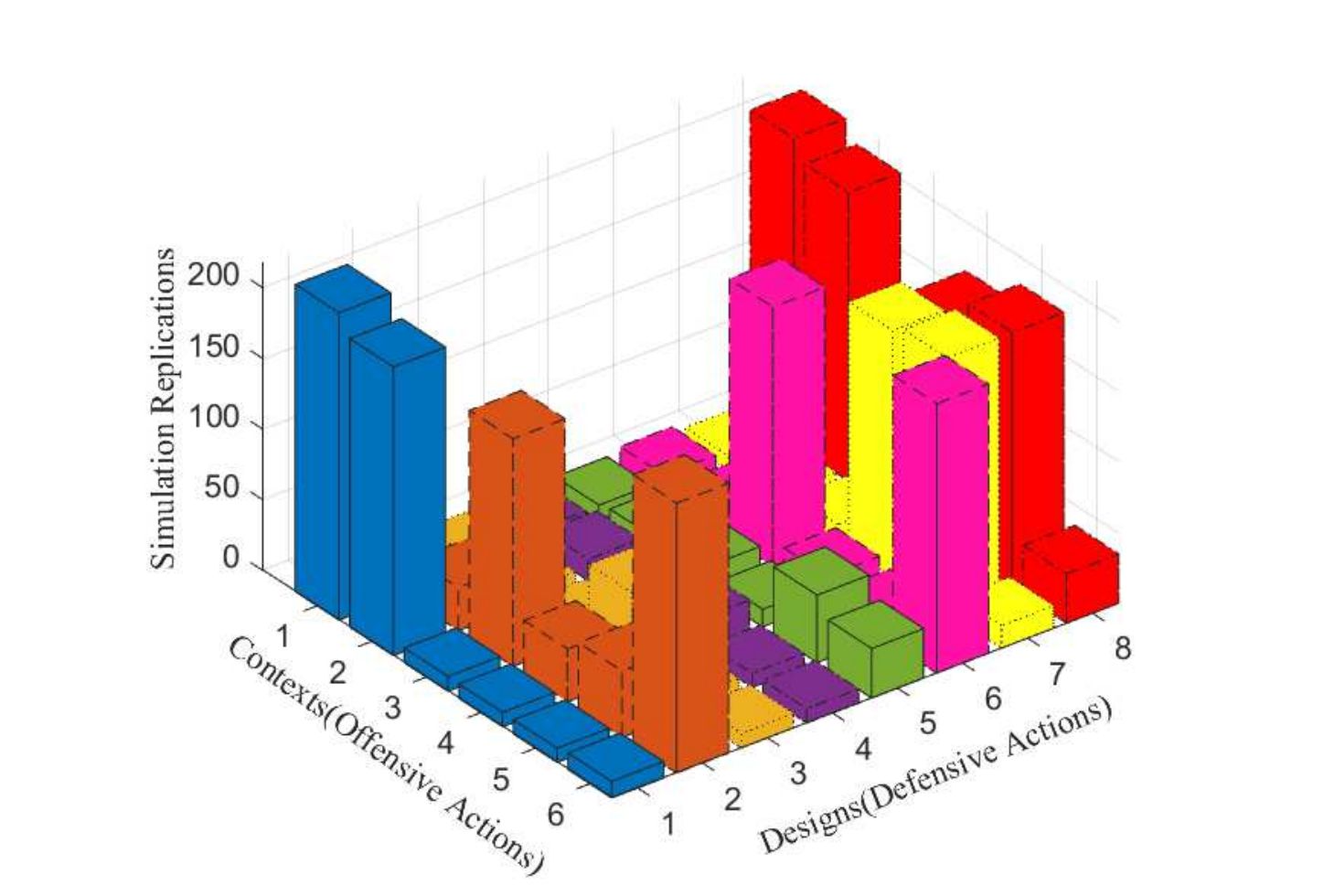}}
\subfigure[m-LinGapE]{\includegraphics[width=0.158\textwidth]{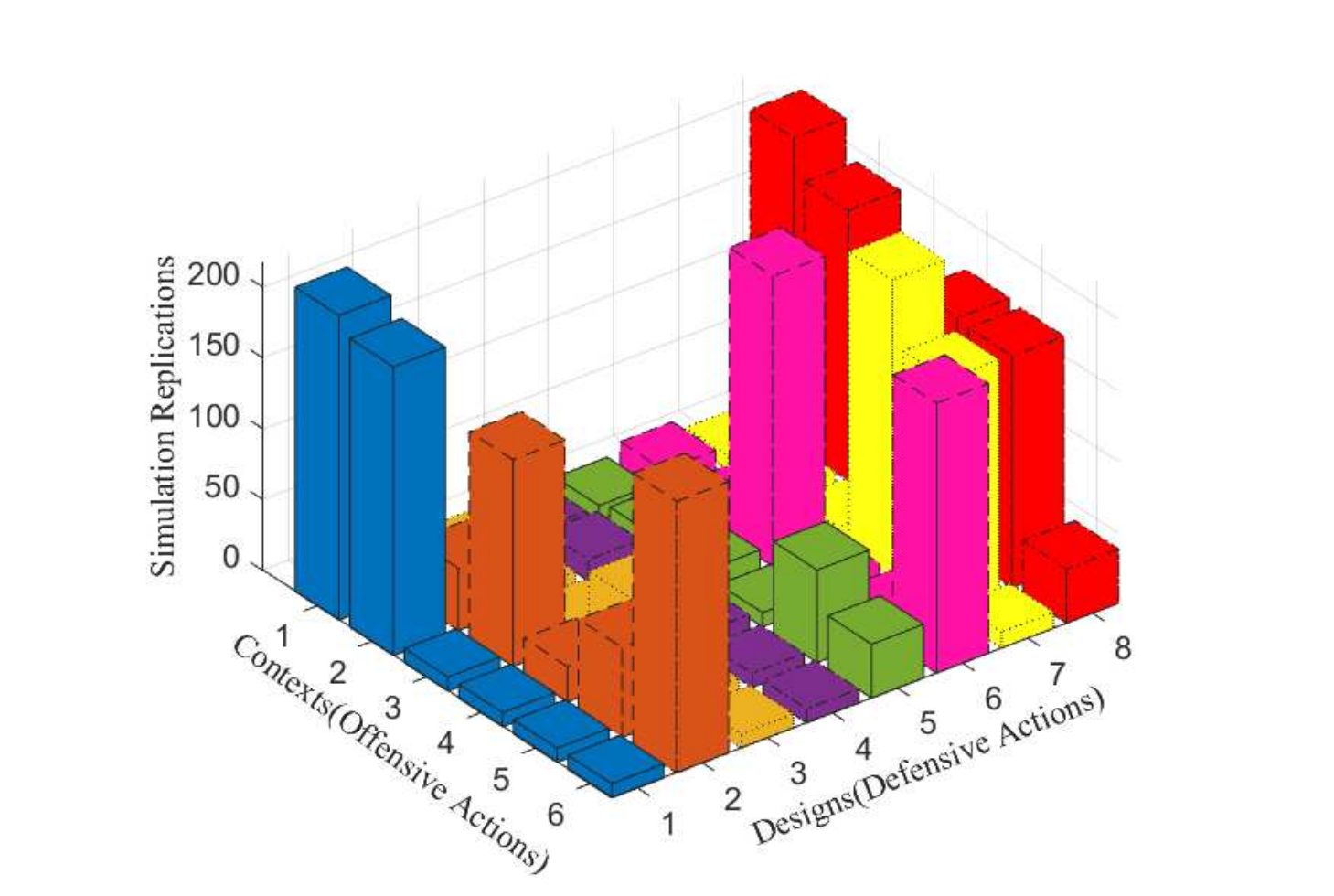}}
\subfigure[BOLDmc]{\includegraphics[width=0.158\textwidth]{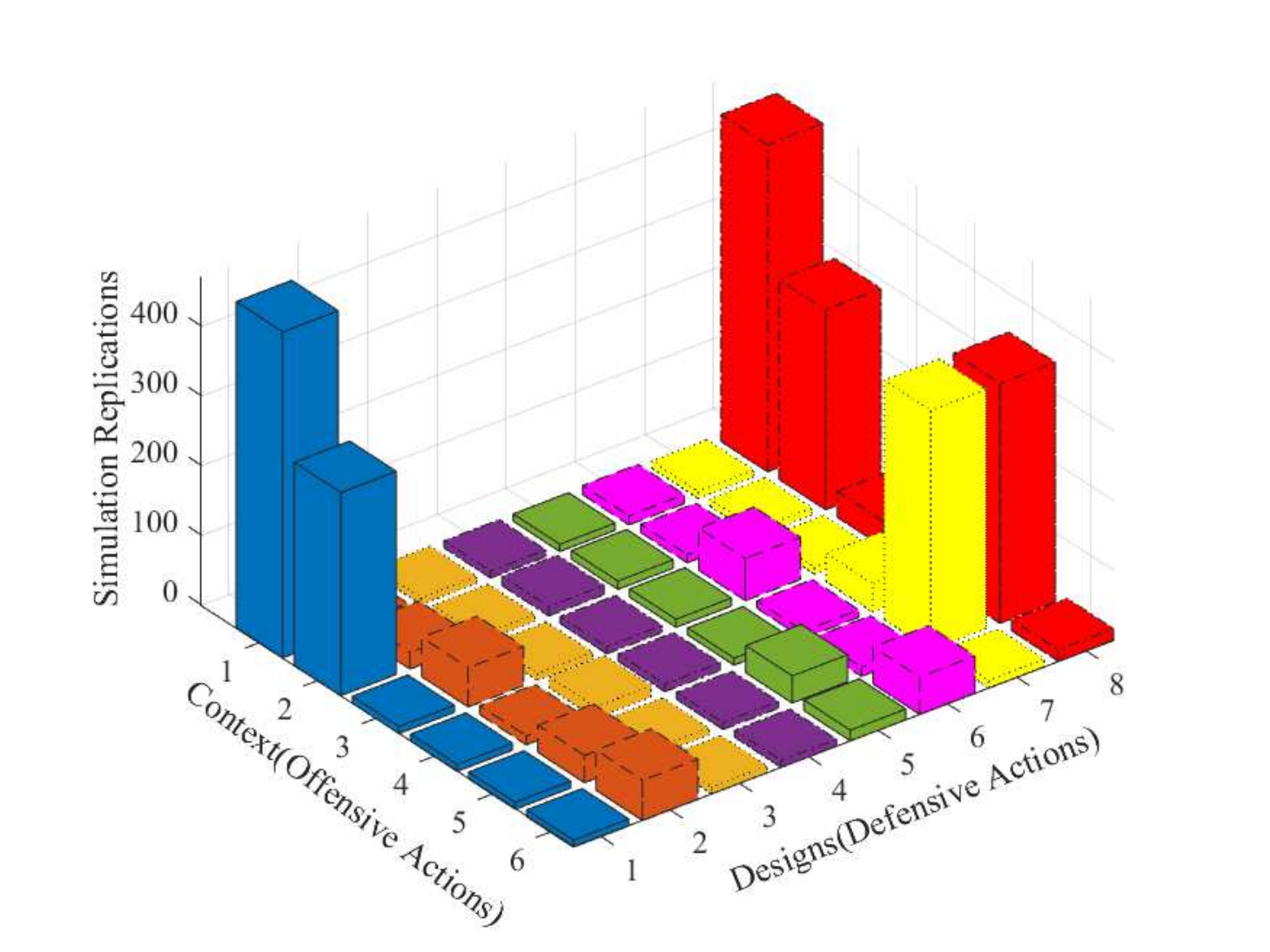}}
\caption{The number of simulation replications allocated to each design-context pair by the sampling policies: (a) AOAmc; (b) EA; (c) E-OCBAm; (d) E-AOAm; (e) m-LinGapE; (f) BOLDmc, in the honeypot deception games example.}
\label{fig8}
\end{figure}

In Figure \ref{fig7}, we can observe that EA performs the worst among all sampling policies. E-AOAm has a slight edge over E-OCBAm at the beginning, and the gap between them narrows as the simulation budget increases. m-LinGapE has a slight edge over E-OCBAm at the beginning, whereas the latter surpasses the former as the simulation budget increases. BOLDmc and AOAmc have a comparable performance at the beginning, whereas the former increases at a slightly slower pace than the latter as the simulation budget increases. AOAmc outperforms all the other compared sampling policies. Figure \ref{fig8} displays the number of simulation replications allocated to each design-context pair by each sampling policy, which is averaged by 100,000 independent macro experiments. From Figure \ref{fig8}, we can see that the design-context pairs receiving more simulation replications are similar across E-OCBAm, E-AOAm, and m-LinGapE. Although the design-context pairs for which AOAmc and BOLDmc allocate more simulation replications are similar, the actual number of allocated simulation replications differs significantly for each design-context pair, leading to differences in ${\rm PCS}_{W}$. This observation coincides with the discussions prior to Experiment 1 regarding the fact that when the prior is uninformative, AOAmc and BOLDmc allocate a simulation replication to the same context at each step, whereas the allocated design could be different. AOAmc allocates more simulation replications to the design-context pairs whose performances are difficult to learn. An efficient sampling procedure for the top-$m$ context-dependent selection problem should consider sampling information among both designs and contexts.

\textit{Experiment 5: medical resource allocation example}. The COVID-19 pandemic has put a tremendous strain on healthcare systems, especially leading to a scarcity of medical resources in emergency departments. In addition to maintaining the normal operations of emergency departments, medical resources are allocated to ensure that admitted patients are free of COVID-19. The main objective of medical resource allocation in emergency departments is to shorten the waiting time for critical patients. Allocation designs in an emergency department are associated with different emergency patient flows, which are highly variable during the COVID-19 pandemic. In this example, we use a discrete-event simulation to evaluate the waiting time for critical patients based on different allocation decisions within each emergency patient flow scenario. Inputs of the simulation model contain design parameters (medical resource allocation decisions) and context parameters (arrival rates).

\begin{figure}[!h]
\centering
\includegraphics[width=0.49\textwidth]{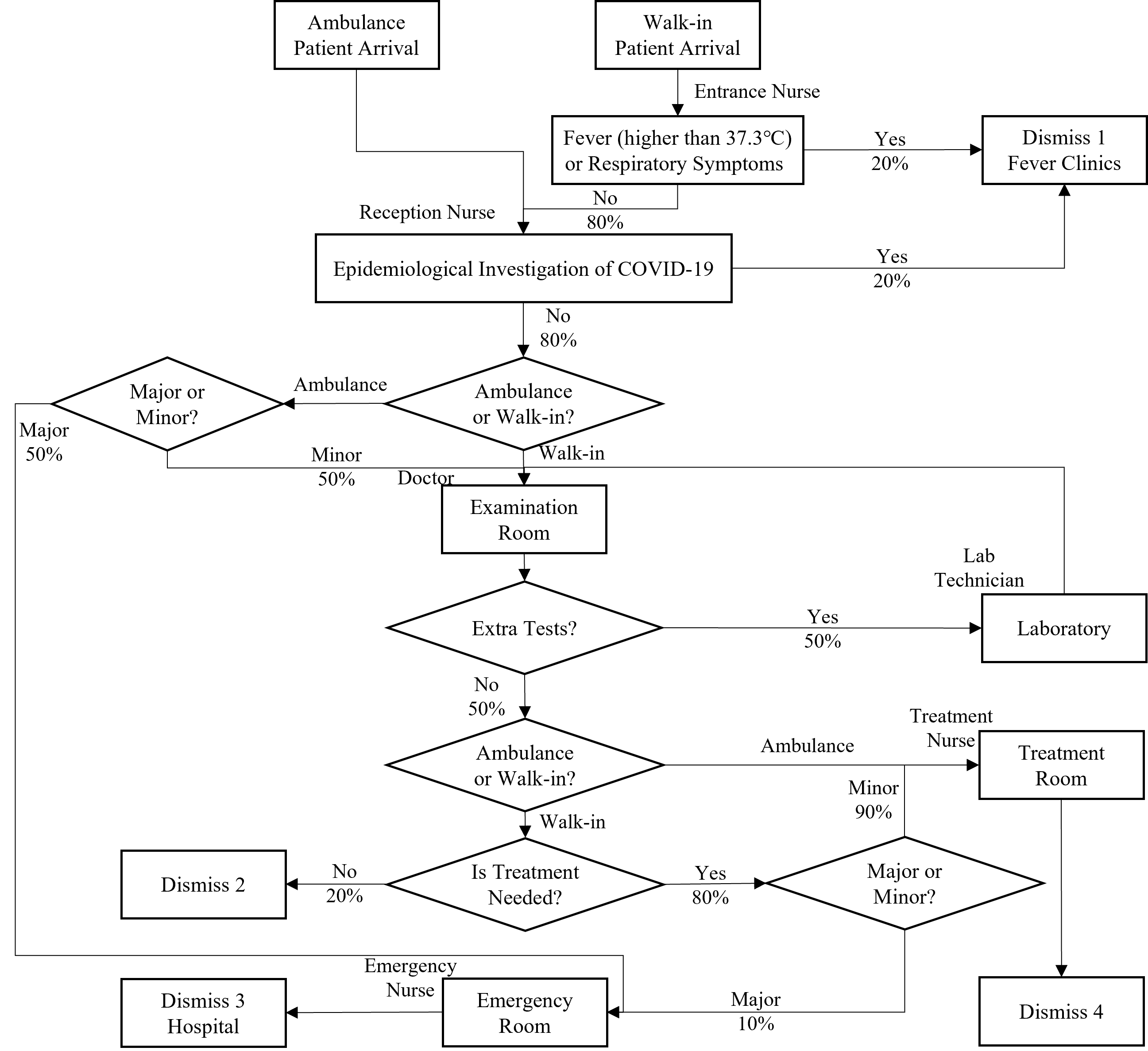}
\caption{Treatment process of patients in an emergency department in Experiment 5.}
\label{fig9}
\end{figure}

Figure \ref{fig9} illustrates the treatment process for emergency patients in an emergency department. Patients who arrive at the emergency department by ambulance are referred to as ambulance patients, whereas other arrivals are categorized as walk-in patients. Walk-in patients go through the entrance, where nurses assess them for COVID-19 symptoms. Those without symptoms go to the reception desk, while ambulance patients rush immediately to the reception desk. Receptionists collect patients' personal information and conduct epidemiological investigations to identify individuals with suspected COVID-19 cases. All COVID-19 suspects identified at the entrance and reception desk are directed to the fever clinics and subsequently discharged from the emergency department. Upon leaving the reception desk, ambulance patients with severe conditions are transported to the emergency room, while doctors in the examination room assess the conditions of other patients. The doctors decide whether patient require additional tests, which are carried out by lab technicians in the laboratory. The tested patients then return to the examination room for further assessment. Subsequently, ambulance patients without severe conditions are transported to the treatment room, while walk-in patients are either discharged from the emergency department after receiving medication or transferred to the treatment room or emergency room based on their conditions. Patients in the treatment room are discharged from the emergency department after finishing their treatments, whereas those in the emergency room are transferred to the hospital for further care before being discharged from the emergency department. The routing probabilities for patients are shown in Figure \ref{fig9}. For the system simulation, we use a 2-day warm-up period and run the simulation for an additional 7 days (24 hours a day). The example is adapted from \cite{bovim2022simulating} and \cite{ahmed2009simulation}.

Table 2 in Section A.4 of the online appendix \cite{zhang2022online} shows the distributions of service times at each stage of the process. Patients are served in the order of their arrival. If no services are available at the current stage, patients remain in the queue until a service becomes available. The arrival process to the emergency department follows a homogeneous Poisson process with rate parameters as shown in Table 3 in Section A.4 of the online appendix \cite{zhang2022online}. The emergency department is equipped with various medical resources, including entrance nurses, receptionists, doctors, lab technicians, treatment beds, and emergency beds. Each medical resource provides services to patients at their corresponding stage of the process. Table 4 in Section A.4 of the online appendix \cite{zhang2022online} presents 20 designs for medical resource allocation. Critical patients refer to those who are routed to the emergency room. The goal is to determine the top-2 designs that minimize the expected waiting time for critical patients in each emergency patient flow scenario. Table 5 in Section A.4 of the online appendix \cite{zhang2022online} displays the performances of the top-4 designs for each emergency patient flow, each estimated by 100,000 independent experiments. We can see that the top-2 designs vary for emergency patient flow, and the differences in their performances are small. The total simulation budget is $T = 5000$.

\begin{figure}[!h]
\centering
\includegraphics[width=0.41\textwidth]{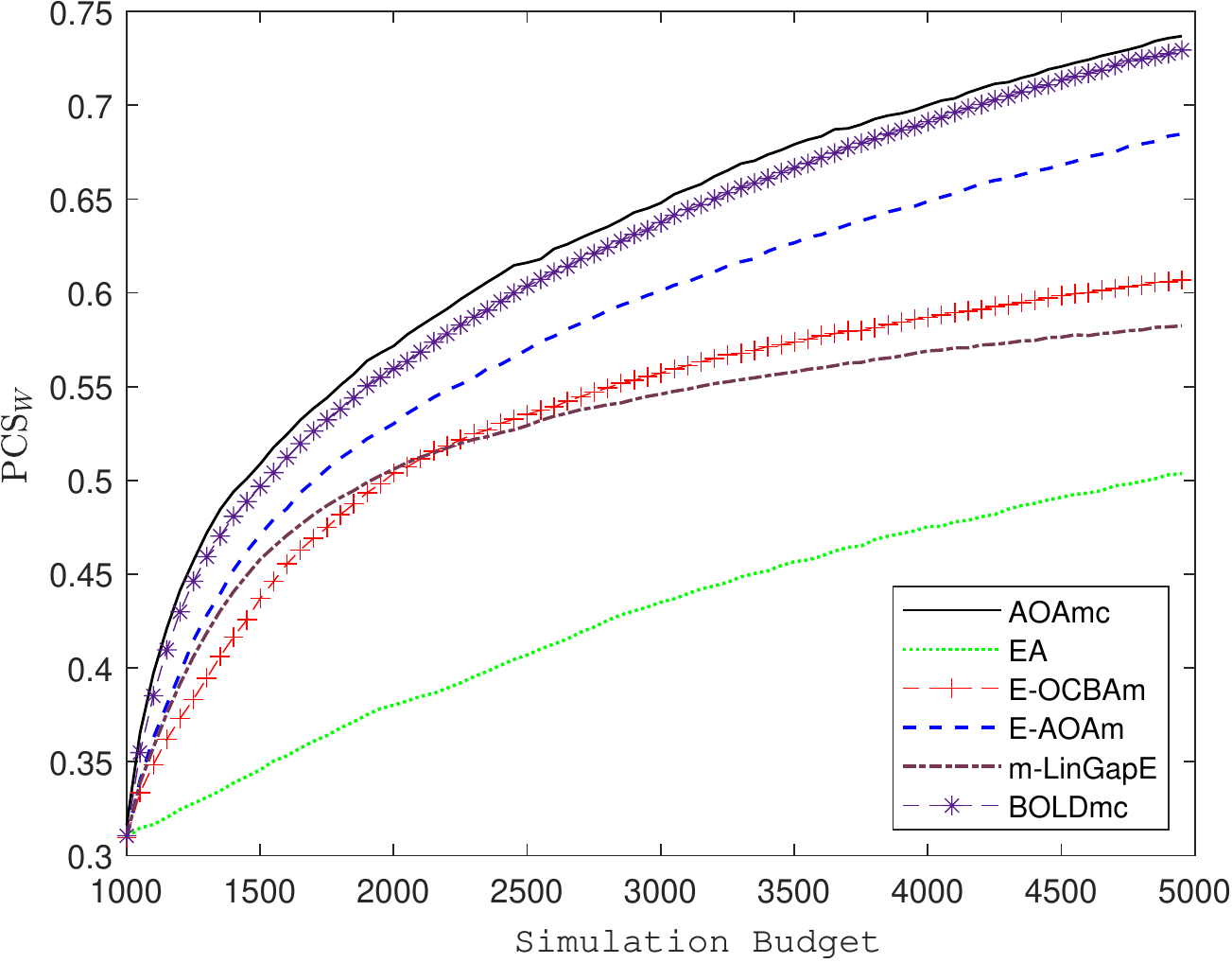}
\caption{${\rm PCS}_{W}$ of the six sampling policies in the medical resource allocation example.}
\label{fig10}
\end{figure}

In Figure \ref{fig10}, we can see that EA performs the worst among all sampling policies. m-LinGapE has a better performance than E-OCBAm at the beginning, whereas the latter surpasses the former as the number of simulation budget grows. E-AOAm has a slight edge over E-OCBAm and m-LinGapE at the beginning, and increases at a faster pace than the latter two sampling policies as the simulation budget increases. The ${\rm PCS}_{W}$ of BOLDmc increases at a slightly slower pace than the ${\rm PCS}_{W}$ of AOAmc. AOAmc performs the best among all the compared sampling policies. AOAmc consumes 4000 simulation replications to attain ${\rm PCS}_{W} = 0.7$, whereas the other sampling policies require more than 4200 simulation replications to achieve the same ${\rm PCS}_{W}$ level, i.e., AOAmc reduces the consumption of simulation replications by more than 4.8\%.

\section{Conclusion}\label{sec6}
The paper studies a simulation budget allocation problem for selecting top-$m$ context-dependent designs. Under a Bayesian framework, we formulate the sequential sampling decision as a stochastic control problem. An efficient sequential sampling procedure named as AOAmc is proposed, which is computationally tractable and is proved to be consistent. To analyze the asymptotic optimality of the proposed sampling policy, the asymptotically optimal sampling ratios which optimize the large deviations rate of the worst-case PFS are rigorously derived. The asymptotic sampling ratios of the proposed allocation policy are shown to be asymptotically optimal under normal sampling distributions. Numerical experiments demonstrate that AOAmc can significantly enhance the efficiency for learning the top-$m$ designs in all contexts. Utilizing parallel computing such that each design-context pair could simultaneously receive multiple simulation replications and exploiting a functional relationship between design and context to enhance the sampling efficiency deserve future research.

\section*{Acknowledgments}
This work was supported in part by the National Natural Science Foundation of China (NSFC) under Grants 72250065, 72022001, 71901003, and the scholarship from China Scholarship Council (CSC) under Grant CSC202206010152. A co-author, Loo Hay Lee, deceased during the writing of this paper. We acknowledge his contribution to this work.

{\appendices
\section*{APPENDIX I: Proof of Corollary \ref{easyimple}}

\begin{IEEEproof}
Notice that if ${A_{t + 1}}\left( {\mathcal{E} _t} \right) = ( {{{\langle {\widetilde i} \rangle }_{rt}},r} )$, ${\widetilde i} \in \left\{1,\cdots,m\right\}$, $r = 1,\cdots,q$, we have ${\left( {\sigma _{{{\left\langle {\widetilde i} \right\rangle }_{r(t + 1)}}}^{\left( {t + 1} \right)}\left( {{{\boldsymbol{x}}_r}} \right)} \right)^2} < {\left( {\sigma _{{{\left\langle {\widetilde i} \right\rangle }_{rt}}}^{\left( t \right)}\left( {{{\boldsymbol{x}}_r}} \right)} \right)^2}$, ${\left( {\sigma _{{{\left\langle \ell \right\rangle }_{r(t + 1)}}}^{\left( {t + 1} \right)}\left( {{{\boldsymbol{x}}_r}} \right)} \right)^2} = {\left( {\sigma _{{{\left\langle \ell \right\rangle }_{rt}}}^{\left( t \right)}\left( {{{\boldsymbol{x}}_r}} \right)} \right)^2}$, $\widetilde {\ell}=1,\cdots,k$, $\widetilde {\ell} \ne {\widetilde i}$,
$$\begin{aligned}
& \mathop {\min }\limits_{j = m + 1, \cdots ,k} \frac{{{{\left( {y_{{{\left\langle {\widetilde i} \right\rangle }_{rt}}}^{\left( t \right)}\left( {{{\boldsymbol{x}}_r}} \right) - y_{{{\left\langle j \right\rangle }_{rt}}}^{\left( t \right)}\left( {{{\boldsymbol{x}}_r}} \right)} \right)}^2}}}{{{{\left( {\sigma _{{{\left\langle {\widetilde i} \right\rangle }_{r(t + 1)}}}^{\left( {t + 1} \right)}\left( {{{\boldsymbol{x}}_r}} \right)} \right)}^2} + {{\left( {\sigma _{{{\left\langle j \right\rangle }_{r(t + 1)}}}^{\left( {t + 1} \right)}\left( {{{\boldsymbol{x}}_r}} \right)} \right)}^2}}} \\
> & \mathop {\min }\limits_{j = m + 1, \cdots ,k} \frac{{{{\left( {y_{{{\left\langle {\widetilde i} \right\rangle }_{rt}}}^{\left( t \right)}\left( {{{\boldsymbol{x}}_r}} \right) - y_{{{\left\langle j \right\rangle }_{rt}}}^{\left( t \right)}\left( {{{\boldsymbol{x}}_r}} \right)} \right)}^2}}}{{{{\left( {\sigma _{{{\left\langle {\widetilde i} \right\rangle }_{rt}}}^{\left( t \right)}\left( {{{\boldsymbol{x}}_r}} \right)} \right)}^2} + {{\left( {\sigma _{{{\left\langle j \right\rangle }_{rt}}}^{\left( t \right)}\left( {{{\boldsymbol{x}}_r}} \right)} \right)}^2}}}~,
\end{aligned}$$
and if ${A_{t + 1}}\left( {{{\mathcal E}_t}} \right) = ( {{{\langle {\widetilde j} \rangle }_{rt}},r} )$, $\widetilde j \in \left\{ {m + 1, \cdots ,k} \right\}$, $r = 1,\cdots,q$, we have ${\left( {\sigma _{{{\left\langle {\widetilde j} \right\rangle }_{r(t + 1)}}}^{\left( {t + 1} \right)}\left( {{{\boldsymbol{x}}_r}} \right)} \right)^2} < {\left( {\sigma _{{{\left\langle {\widetilde j} \right\rangle }_{rt}}}^{\left( t \right)}\left( {{{\boldsymbol{x}}_r}} \right)} \right)^2}$, ${\left( {\sigma _{{{\left\langle \widetilde{\ell} \right\rangle }_{r(t + 1)}}}^{\left( {t + 1} \right)}\left( {{{\boldsymbol{x}}_r}} \right)} \right)^2} = {\left( {\sigma _{{{\left\langle \widetilde{\ell} \right\rangle }_{rt}}}^{\left( t \right)}\left( {{{\boldsymbol{x}}_r}} \right)} \right)^2}$, $\widetilde{\ell}=1,\cdots,k$, $\widetilde{\ell} \ne {\widetilde j}$,
$$\begin{aligned}
& \mathop {\min }\limits_{i = 1, \cdots ,m} \frac{{{{\left( {y_{{{\left\langle i \right\rangle }_{rt}}}^{\left( t \right)}\left( {{{\boldsymbol{x}}_r}} \right) - y_{{{\left\langle {\widetilde j} \right\rangle }_{rt}}}^{\left( t \right)}\left( {{{\boldsymbol{x}}_r}} \right)} \right)}^2}}}{{{{\left( {\sigma _{{{\left\langle i \right\rangle }_{r(t + 1)}}}^{\left( {t + 1} \right)}\left( {{{\boldsymbol{x}}_r}} \right)} \right)}^2} + {{\left( {\sigma _{{{\left\langle {\widetilde j} \right\rangle }_{r(t + 1)}}}^{\left( {t + 1} \right)}\left( {{{\boldsymbol{x}}_r}} \right)} \right)}^2}}} \\
> & \mathop {\min }\limits_{i = 1, \cdots ,m} \frac{{{{\left( {y_{{{\left\langle i \right\rangle }_{rt}}}^{\left( t \right)}\left( {{{\boldsymbol{x}}_r}} \right) - y_{{{\left\langle {\widetilde j} \right\rangle }_{rt}}}^{\left( t \right)}\left( {{{\boldsymbol{x}}_r}} \right)} \right)}^2}}}{{{{\left( {\sigma _{{{\left\langle i \right\rangle }_{rt}}}^{\left( t \right)}\left( {{{\boldsymbol{x}}_r}} \right)} \right)}^2} + {{\left( {\sigma _{{{\left\langle {\widetilde j} \right\rangle }_{rt}}}^{\left( t \right)}\left( {{{\boldsymbol{x}}_r}} \right)} \right)}^2}}}~.
\end{aligned}$$

Let ${\widetilde {\ell}^{\left( t + 1 \right)}} \notin \arg \mathop {\min }\nolimits_{\ell = 1, \cdots ,q} {APCS_\ell}\left( {{\mathcal{E}_t}} \right)$, ${\widehat {\ell}^{\left( t + 1 \right)}} \in \arg \mathop {\min }\nolimits_{\ell = 1, \cdots ,q} {APCS_\ell}\left( {{\mathcal{E}_t}} \right)$, i.e., ${APCS_{\widetilde {\ell}^{\left( t + 1\right)}}}\left( {{\mathcal{E}_t}} \right) > {APCS_{\widehat {\ell}^{\left( t + 1 \right)}}}\left( {{\mathcal{E}_t}} \right)$. Suppose that following the sampling rule (\ref{AOAmc}), ${A_{t + 1}}\left( {{{\cal E}_t}} \right) = ( {\widetilde h,{{\widetilde \ell }^{\left( t + 1 \right)}}} ),\widetilde h \in \left\{ {1, \cdots ,k} \right\}$, and then we have
$$\begin{aligned}
& {APCS_{{{\widetilde \ell }^{\left( t + 1 \right)}}}}\left( {{\mathcal E}_{t + 1}^E} \right) = {APCS_{{{\widetilde \ell }^{\left( t + 1 \right)}}}}\left( {{\mathcal E}_t} \cup \mathbb{E}\left[ {\left. {Y_{\widetilde h{{\widetilde \ell }^{\left( t + 1\right)}}}^{{t_{\widetilde h{{\widetilde \ell }^{\left( t + 1\right)}}}} + 1}} \right|{{\mathcal E}_t}} \right] \right) \\
& \ge {APCS_{{{\widetilde \ell }^{\left( t + 1\right)}}}}\left( {{{\mathcal E}_t}} \right) > {APCS_{{{\widehat \ell }^{\left( t + 1\right)}}}}\left( {{{\mathcal E}_t}} \right)~,
\end{aligned}$$
leading to, for $\widetilde \ell  = 1, \cdots ,k$, $\forall {{\widetilde \ell }^{\left( t + 1\right)}},{{\widehat \ell }^{\left( t + 1\right)}}$,
$$\begin{aligned}
{{\widehat V}_t}\left( {{{\mathcal E}_t};\left( {\widetilde \ell ,{{\widetilde \ell }^{\left( t + 1 \right)}}} \right)} \right) & = \mathop {\min }\limits_{\ell  = 1, \cdots ,q} {{APCS}_\ell } \left( {{\mathcal E}_t} \cup \mathbb{E} [ { {Y_{\widetilde \ell{{\widetilde \ell }^{\left( t + 1 \right)}}}^{{t_{\widetilde \ell{{\widetilde \ell }^{\left( t + 1 \right)}}}} + 1}} |{{\mathcal E}_t}} ] \right) \\
& = {APCS_{{{\widehat \ell }^{\left( t + 1 \right)}}}}\left( {{{\mathcal E}_{t}}} \right) = \mathop {\min }\limits_{\ell  = 1, \cdots ,q} {{APCS}_\ell }\left( {{{\mathcal E}_t}} \right),
\end{aligned}$$
which contradicts the sampling rule (\ref{AOAmc}) that the allocated design-context pair  has the maximal ${{\widehat V}_t}( {{{\mathcal E}_t};( {\widetilde \ell ,r} )} )$, ${\widetilde \ell} = 1,\cdots,k$, $r = 1,\cdots,q$. Therefore, the allocated context of the design-context pair following the sampling rule (\ref{AOAmc}) is ${\widehat {\ell}^{\left( t + 1 \right)}} \in \arg \mathop {\min }\limits_{\ell = 1, \cdots ,q} {APCS_\ell}\left( {{\mathcal{E}_t}} \right)$.

Let set ${{\widetilde H}^{\left( t + 1\right)}} = {{\left\{ {1, \cdots ,k} \right\}} \setminus {{H^{\left( t + 1 \right)}}}}$, where $ B \setminus A$ denotes relative complement of set A in set B. Suppose that following the sampling rule (\ref{AOAmc}), ${A_{t + 1}}\left( {{\mathcal{E}_t}} \right) = ( {{\widetilde h}^{\left( t + 1 \right)}},{{\widetilde \ell }^{\left( t + 1 \right)}})$, $\forall~{{\widetilde h}^{\left( t + 1 \right)}} \in {\widetilde H}^{\left( t + 1\right)}$, and then we have
$${APCS_{{{\widehat \ell }^{\left( t + 1\right)}}}}\left( {{{\mathcal E}_t} \cup \mathbb{E} [ { {Y_{{{\widetilde h}^{\left( t + 1 \right)}}{{\widetilde \ell }^{\left( t + 1 \right)}}}^{{t_{{{\widetilde h}^{\left( t + 1 \right)}}{{\widetilde \ell }^{\left( t + 1\right)}}}} + 1}} |{{\mathcal E}_t}} ]} \right) = {APCS_{{{\widehat \ell }^{\left( t + 1\right)}}}}\left( {{{\mathcal E}_t}} \right),$$
leading to, $\forall~{{\widetilde h}^{\left( t + 1\right)}}$,
$$\begin{aligned}
& {{\widehat V}_t}\left( {{{\mathcal E}_t};\left( {{{\widetilde h}^{\left( t + 1\right)}},{{\widetilde \ell }^{\left( t + 1\right)}}} \right)} \right){\rm{ }} \\
= & \mathop {\min }\limits_{\ell  = 1, \cdots ,q} APC{S_\ell }\left( {{{\mathcal E}_t} \cup \mathbb{E}\left[ {\left. {Y_{{{\widetilde h}^{\left( t + 1 \right)}}{{\widetilde \ell }^{\left( t + 1 \right)}}}^{{t_{{{\widetilde h}^{\left( t + 1 \right)}}{{\widetilde \ell }^{\left( t + 1\right)}}}} + 1}} \right|{{\mathcal E}_t}} \right]} \right) \\
= & {APCS_{{{\widehat \ell }^{\left( t + 1 \right)}}}}\left( {{{\mathcal E}_t}} \right) = \mathop {\min }\limits_{\ell  = 1, \cdots ,q} {APCS_\ell }\left( {{{\mathcal E}_t}} \right)~,
\end{aligned}$$
which contradicts the sampling rule (\ref{AOAmc}) that the allocated design-context pair  has the maximal ${{\widehat V}_t}( {{{\mathcal E}_t};( {\widetilde \ell ,r} )} )$, ${\widetilde \ell} = 1,\cdots,k$, $r = 1,\cdots,q$. Therefore, the allocated design ${{\widehat i}^{\left( t + 1\right)}}$ under the context ${\widetilde \ell }^{\left( t + 1 \right)}$ following the sampling rule (\ref{AOAmc}) belongs to the set ${H^{\left( t + 1\right)}}$, and can be calculated as, $\forall~{h^{\left( t + 1\right)}} \in {H^{\left( t + 1 \right)}}$,
$${{\widehat i}^{\left( t + 1\right)}} \in \arg \mathop {\max }\limits_{{{{\langle h^{\left(t + 1\right)} \rangle }_{{{\widehat \ell }^{\left( t + 1\right)}}t}}}} {{\widehat V}_t} \left( {{\mathcal{E} _t}; \left( {{{{\left\langle h^{\left(t + 1\right)} \right\rangle }_{{{\widehat \ell }^{\left( t + 1\right)}}t}}},{{\widehat \ell }^{\left( t + 1\right)}}} \right)} \right)~.$$

Summarizing the above, the Corollary is proved.
\end{IEEEproof}

\section*{APPENDIX II: Proof of Corollary \ref{saratocon}}

\begin{IEEEproof}
Following (\ref{balance1}), for $\ell \in \left\{1,\cdots,q\right\}$, we have
$$\begin{aligned}
& \mathop {\min }\limits_{j = m + 1, \cdots ,k} \frac{{{{r_\ell^*}\left( {{y _{{{\left\langle i \right\rangle }_\ell}}\left(\boldsymbol{x}_{\ell}\right)} - {y _{{{\left\langle j \right\rangle }_\ell}}\left(\boldsymbol{x}_{\ell}\right)}} \right)^2}}}{{{{\sigma _{{{\left\langle i \right\rangle }_\ell }}^2\left( {{{\bf{x}}_\ell }} \right)} \mathord{\left/{\vphantom {{\sigma _{{{\left\langle i \right\rangle }_\ell }}^2\left( {{{\boldsymbol{x}}_\ell }} \right)} {\left( {{{r_{\left\langle i \right\rangle \ell }^*} \mathord{\left/{\vphantom {{r_{\left\langle i \right\rangle \ell }^*} {r_\ell ^*}}} \right.\kern-\nulldelimiterspace} {r_\ell ^*}}} \right)}}} \right.\kern-\nulldelimiterspace} {( {{{r_{\left\langle i \right\rangle \ell }^*} / {r_\ell ^*}}} )}} + {{{\sigma _{{{\left\langle j \right\rangle }_\ell }}^2\left( {{{\boldsymbol{x}}_\ell }} \right)} \mathord{\left/
 {\vphantom {{\sigma _{{{\left\langle j \right\rangle }_\ell }}^2\left( {{{\bf{x}}_\ell }} \right)} {\left( {{{r_{\left\langle j \right\rangle \ell }^*} \mathord{\left/{\vphantom {{r_{\left\langle j \right\rangle \ell }^*} {r_\ell ^*}}} \right.\kern-\nulldelimiterspace} {r_\ell ^*}}} \right)}}} \right.\kern-\nulldelimiterspace} {( {{{r_{\left\langle j \right\rangle \ell }^*} / {r_\ell ^*}}} )}}}}} \\
 = & \mathop {\min }\limits_{i' = 1, \cdots ,m} \frac{{{{{r_\ell^*} \left( {{y _{{{\left\langle {i'} \right\rangle }_\ell}}\left(\boldsymbol{x}_{\ell}\right)} - {y _{{{\left\langle {j'} \right\rangle }_\ell}}\left(\boldsymbol{x}_{\ell}\right)}} \right)}^2}}}{{{{\sigma _{{{\left\langle {i'} \right\rangle }_\ell }}^2\left( {{{\boldsymbol{x}}_\ell }} \right)} \mathord{\left/
 {\vphantom {{\sigma _{{{\left\langle {i'} \right\rangle }_\ell }}^2\left( {{{\bf{x}}_\ell }} \right)} {\left( {{{r_{\left\langle {i'} \right\rangle \ell }^*} \mathord{\left/{\vphantom {{r_{\left\langle {i'} \right\rangle \ell }^*} {r_\ell ^*}}} \right.\kern-\nulldelimiterspace} {r_\ell ^*}}} \right)}}} \right.\kern-\nulldelimiterspace} {( {{{r_{\left\langle {i'} \right\rangle \ell }^*} / {r_\ell ^*}}} )}} + {{{\sigma _{{{\left\langle {j'} \right\rangle }_\ell }}^2\left( {{{\bf{x}}_\ell }} \right)} \mathord{\left/
 {\vphantom {{\sigma _{{{\left\langle {j'} \right\rangle }_\ell }}^2\left( {{{\bf{x}}_\ell }} \right)} {\left( {{{r_{\left\langle {j'} \right\rangle \ell }^*} \mathord{\left/{\vphantom {{r_{\left\langle {j'} \right\rangle \ell }^*} {r_\ell ^*}}} \right.\kern-\nulldelimiterspace} {r_\ell ^*}}} \right)}}} \right.\kern-\nulldelimiterspace} {( {{{r_{\left\langle {j'} \right\rangle \ell }^*} / {r_\ell ^*}}} )}}}}}~,
 \end{aligned}$$
 and then (\ref{corsingequ}) holds by letting $\alpha _{\left\langle h \right\rangle \ell}^* = {{r_{\left\langle h \right\rangle \ell}^*} \mathord{\left/{\vphantom {{r_{\left\langle h \right\rangle \ell}^*} {r_\ell^*}}} \right.\kern-\nulldelimiterspace} {r_\ell^*}}$, $h = 1,\cdots,k$. With ${r_{\left\langle h \right\rangle \ell}^*} > 0$, and $\sum\nolimits_{h = 1}^k {r_{\left\langle h \right\rangle \ell}^*}  = r_\ell^* > 0$, we have $\boldsymbol{\alpha} _\ell^* > 0$, and $\sum\nolimits_{h = 1}^k {\alpha _{h\ell}^*}  = 1$. In addition, (\ref{balance1}) also leads to, for $\ell,\ell^{\prime} = 1,\cdots,q$,
 $$\begin{aligned}
& \mathop {\min }\limits_{j = m + 1, \cdots ,k} \frac{{{{r_\ell^*}\left( {{y _{{{\left\langle i \right\rangle }_\ell}}\left(\boldsymbol{x}_{\ell}\right)} - {y _{{{\left\langle j \right\rangle }_\ell}}\left(\boldsymbol{x}_{\ell}\right)}} \right)^2}}}{{{{\sigma _{{{\left\langle i \right\rangle }_\ell }}^2\left( {{{\bf{x}}_\ell }} \right)} \mathord{\left/{\vphantom {{\sigma _{{{\left\langle i \right\rangle }_\ell }}^2\left( {{{\boldsymbol{x}}_\ell }} \right)} {\left( {{{r_{\left\langle i \right\rangle \ell }^*} \mathord{\left/{\vphantom {{r_{\left\langle i \right\rangle \ell }^*} {r_\ell ^*}}} \right.\kern-\nulldelimiterspace} {r_\ell ^*}}} \right)}}} \right.\kern-\nulldelimiterspace} {( {{{r_{\left\langle i \right\rangle \ell }^*} / {r_\ell ^*}}} )}} + {{{\sigma _{{{\left\langle j \right\rangle }_\ell }}^2\left( {{{\boldsymbol{x}}_\ell }} \right)} \mathord{\left/
 {\vphantom {{\sigma _{{{\left\langle j \right\rangle }_\ell }}^2\left( {{{\bf{x}}_\ell }} \right)} {\left( {{{r_{\left\langle j \right\rangle \ell }^*} \mathord{\left/{\vphantom {{r_{\left\langle j \right\rangle \ell }^*} {r_\ell ^*}}} \right.\kern-\nulldelimiterspace} {r_\ell ^*}}} \right)}}} \right.\kern-\nulldelimiterspace} {( {{{r_{\left\langle j \right\rangle \ell }^*} / {r_\ell ^*}}} )}}}}} = \\
 & \mathop {\min }\limits_{j = m + 1, \cdots ,k} \frac{{{{r_{\ell^{\prime}}^*}\left( {{y _{{{\left\langle i \right\rangle }_{\ell^{\prime}}}}\left(\boldsymbol{x}_{{\ell^{\prime}}}\right)} - {y _{{{\left\langle j \right\rangle }_{\ell^{\prime}}}}\left(\boldsymbol{x}_{{\ell^{\prime}}}\right)}} \right)^2}}}{{{{\sigma _{{{\left\langle i \right\rangle }_{\ell^{\prime}} }}^2\left( {{{\bf{x}}_{\ell^{\prime}} }} \right)} \mathord{\left/{\vphantom {{\sigma _{{{\left\langle i \right\rangle }_{\ell^{\prime}} }}^2\left( {{{\boldsymbol{x}}_{\ell^{\prime}} }} \right)} {\left( {{{r_{\left\langle i \right\rangle {\ell^{\prime}} }^*} \mathord{\left/{\vphantom {{r_{\left\langle i \right\rangle {\ell^{\prime}} }^*} {r_{\ell^{\prime}} ^*}}} \right.\kern-\nulldelimiterspace} {r_{\ell^{\prime}} ^*}}} \right)}}} \right.\kern-\nulldelimiterspace} {( {{{r_{\left\langle i \right\rangle {\ell^{\prime}} }^*} / {r_{\ell^{\prime}} ^*}}} )}} + {{{\sigma _{{{\left\langle j \right\rangle }_{\ell^{\prime}} }}^2\left( {{{\boldsymbol{x}}_{\ell^{\prime}} }} \right)} \mathord{\left/
 {\vphantom {{\sigma _{{{\left\langle j \right\rangle }_{\ell^{\prime}} }}^2\left( {{{\bf{x}}_{\ell^{\prime}} }} \right)} {\left( {{{r_{\left\langle j \right\rangle {\ell^{\prime}} }^*} / {r_{\ell^{\prime}} ^*}}} \right)}}} \right.\kern-\nulldelimiterspace} {( {{{r_{\left\langle j \right\rangle {\ell^{\prime}} }^*} / {r_{\ell^{\prime}} ^*}}} )}}}}},
 \end{aligned}$$
Following the definition of ${z_\ell }\left( \cdot \right)$, we have $r_\ell ^*{z_\ell }\left( {\alpha _\ell ^*} \right) = r_{{\ell ^\prime }}^*{z_{{\ell ^\prime }}}\left( {\alpha _{{\ell ^\prime }}^*} \right)$. With $r_{\ell}^* > 0$, and $\sum\nolimits_{\ell = 1}^q {r_\ell^*}  = 1$, it leads to
$$r_\ell^* = \frac{{{1 \mathord{\left/
{\vphantom {1 {{z_\ell}\left( {\boldsymbol{\alpha}_\ell^*} \right)}}} \right.\kern-\nulldelimiterspace} {{z_\ell}\left( {\boldsymbol{\alpha}_\ell^*} \right)}}}}{{\sum\nolimits_{\widetilde \ell = 1}^q {{1 \mathord{\left/{\vphantom {1 {{z_{\widetilde \ell}}\left( {\alpha _{\widetilde \ell}^*} \right)}}} \right.\kern-\nulldelimiterspace} {{z_{\widetilde \ell}}\left( {\alpha _{\widetilde \ell}^*} \right)}}} }},\quad \ell=1,\cdots,q~.$$
According to the definition of $z^*$, and under the normality assumption, $z^*$ can be expressed as: $z^* = \mathop {\min }\limits_{j = m + 1, \cdots ,k} \frac{{{\left( {{y _{{{\left\langle i \right\rangle }_\ell}}\left(\boldsymbol{x}_{\ell}\right)} - {y _{{{\left\langle j \right\rangle }_\ell}}\left(\boldsymbol{x}_{\ell}\right)}} \right)^2}}}{{{{\sigma _{{{\left\langle i \right\rangle }_\ell}}^2}\left(\boldsymbol{x}_{\ell}\right) \mathord{\left/
 {\vphantom {{\sigma _{{{\left\langle i \right\rangle }_\ell}}^2} {r_{i\ell}^*}}} \right.\kern-\nulldelimiterspace} {r_{{\left\langle i \right\rangle }\ell}^*}} + {{\sigma _{{{\left\langle j \right\rangle}_\ell}}^2}\left(\boldsymbol{x}_{\ell}\right) \mathord{\left/{\vphantom {{\sigma _{{{\left\langle j \right\rangle}_\ell}}^2} {r_{j\ell}^*}}} \right.\kern-\nulldelimiterspace} {r_{{\left\langle j \right\rangle }\ell}^*}}}}$, leading to
 $${z^*} = r_\ell ^*{z_\ell }\left( {\alpha _\ell ^*} \right) = \frac{1}{{\sum\nolimits_{\widetilde \ell = 1}^q {{1 \mathord{\left/{\vphantom {1 {{z_{\widetilde \ell}}\left( {\alpha _{\widetilde \ell}^*} \right)}}} \right.\kern-\nulldelimiterspace} {{z_{\widetilde \ell}}\left( {\alpha _{\widetilde \ell}^*} \right)}}} }}~.$$
\end{IEEEproof}
}


%

\bibliographystyle{IEEEtran}
\bibliography{IEEEabrv,IEEEexample}

\vfill

\end{document}